\documentclass{article}

\usepackage{arabtex}
\usepackage{utf8} 
\setcode{utf8}
\usepackage[final]{neurips_data_2024}
\usepackage[utf8]{inputenc} 
\usepackage[T1]{fontenc}    
\usepackage{hyperref}       
\usepackage{url}            
\usepackage{booktabs}       
\usepackage{amsfonts}       
\usepackage{nicefrac}       
\usepackage{microtype}      
\usepackage{xcolor}         
\usepackage{subcaption}
\usepackage{graphicx}
\usepackage{csvsimple,longtable} 
\usepackage{array} 
\usepackage{verbatim}
\usepackage{markdown}
\usepackage{pifont}
\usepackage[group-separator={,},group-minimum-digits=4,separate-uncertainty=true]{siunitx}
\usepackage{caption}
\usepackage{multirow}
\newcommand{\changecolor}{}
\newcommand{\magentaText}{}
\newcommand{\blueText}{}

\title{Muharaf: Manuscripts of Handwritten Arabic Dataset for Cursive Text Recognition}

\author{%
Mehreen Saeed$^{1}$\quad Adrian Chan$^{1}$\quad Anupam Mijar$^{1}$\quad Joseph Moukarzel$^{2}$\\ \quad \textbf{Georges Habchi}$^{2}$ \quad \textbf{Carlos Younes}$^{2}$ \quad \textbf{Amin Elias}$^{3}$ \quad \textbf{Chau-Wai Wong}$^{1}$ \quad \textbf{Akram Khater}$^{1}$\\$^{1}$North Carolina State University, \quad $^{2}$Holy Spirit University of Kaslik \\$^{3}$Lebanese Association for History\\
\texttt{\{mehreen.mehreen,adrian27513,aamijar230\}@gmail.com}\\\texttt{  \{josephmoukarzel,georgeshabchi,carlosyounes\}@usek.edu.lb}\\
\texttt{  a.elias@lahlebanon.org}\\
\texttt{\{chauwai.wong,akhater\}@ncsu.edu}\\
}

\begin{document}
\maketitle

\begin{abstract}
We present the Manuscripts of Handwritten Arabic~(Muharaf) dataset, which is a machine learning dataset consisting of more than 1,600 historic handwritten page images transcribed by experts in archival Arabic. Each document image is accompanied by spatial polygonal coordinates of its text lines as well as basic page elements. This dataset was compiled to advance the state of the art in handwritten text recognition (HTR), not only for Arabic manuscripts but also for cursive text in general. The Muharaf dataset includes diverse handwriting styles and a wide range of document types, including personal letters, diaries, notes, poems, church records, and legal correspondences. In this paper, we describe the data acquisition pipeline, notable dataset features, and statistics. We also provide a preliminary baseline result achieved by training convolutional neural networks using this data. 
\end{abstract}

\section{Introduction}
Modern standard Arabic has more than 400 million native speakers worldwide and is the official language of 24 sovereign countries as of 2024~\cite{arabic_wikipedia}. Arabic is not only widely spoken but also has a vast collection of historical manuscripts spanning rich literary traditions, poetry, philosophy, and scientific writings. The British Library alone has a massive collection of almost \num{15000} works in \num{14000} volumes of Arabic manuscripts~\cite{rasam_2021}. \changecolor{A highly accurate optical character recognition (OCR) system for handwritten historic Arabic manuscripts will make these documents accessible to a global community of researchers, historians, literary scholars, linguists, and genealogists.}  

\changecolor{In the past decade, handwritten text recognition (HTR) has made significant progress through the use of deep neural networks~\cite{sfr_2018, dan_2020, metahtr_2021, trocr_2021, adocrnet_2024}. Unlike traditional HTR systems that employ handcrafted features, these networks are data-hungry and require significant amounts of training data to learn, generalize, and be deployed in real-world scenarios. For Arabic HTR, there are unique challenges involved.} The Arabic script is cursive and involves varying letter shapes depending on their positions within a word. Moreover, the harakat and diacritics of the Arabic script add to the difficulty of the task. The scarcity of public datasets, compounded by their relatively small sizes, further exacerbates the challenges.

\changecolor{We created the \underline Man\underline uscripts of \underline{Ha}ndwritten \underline{Ara}bic (Muharaf\footnote{``Muharaf'' is Arabic for ``typeface''.}) dataset of fully annotated and transcribed \num{1644} images to train and evaluate an HTR system for Arabic handwritten historical manuscripts.} The document images for this dataset were acquired from the archives of the Phoenix Center for Lebanese Studies at Holy Spirit University of Kaslik (USEK) and the Khayrallah Center for Lebanese Diaspora Studies (KCLDS) at North Carolina State University (NC State). In the preliminary phase of the data collection pipeline, experts in archival Arabic annotated and transcribed the individual text lines in the document images. In the main phase, we leveraged deep learning to predict the texts of the document pages, which was then manually corrected by experts. 

The Muharaf dataset can be used not only in HTR systems but also in other text-related tasks such as text-line segmentation, layout analysis, writer identification, style classification, and more. While Muharaf is not customized for training a page layout detection system, it contains the annotation and labeling of some basic page elements. For example, graphics, page numbers, floating regions, crossed-out text, paragraph separators, and signature areas have all been marked and annotated. 

\begin{figure*}
\centering
\begin{subfigure}[t]{0.195\textwidth}
 \includegraphics[width=\linewidth, height=3cm]{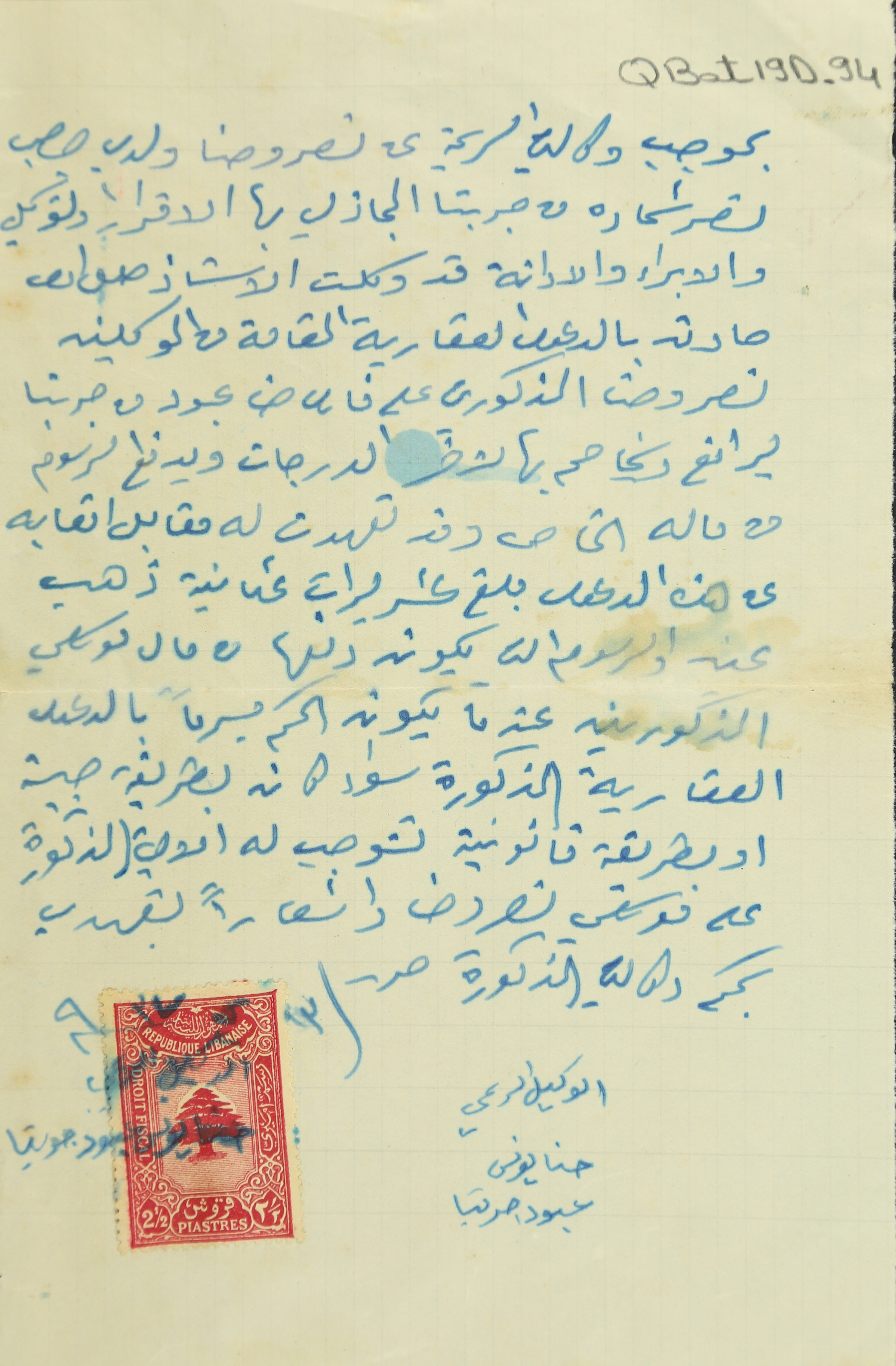}
    \caption{}\label{QBat}
  \end{subfigure}
\begin{subfigure}[t]{0.195\textwidth}
 \includegraphics[width=\linewidth, height=3cm]{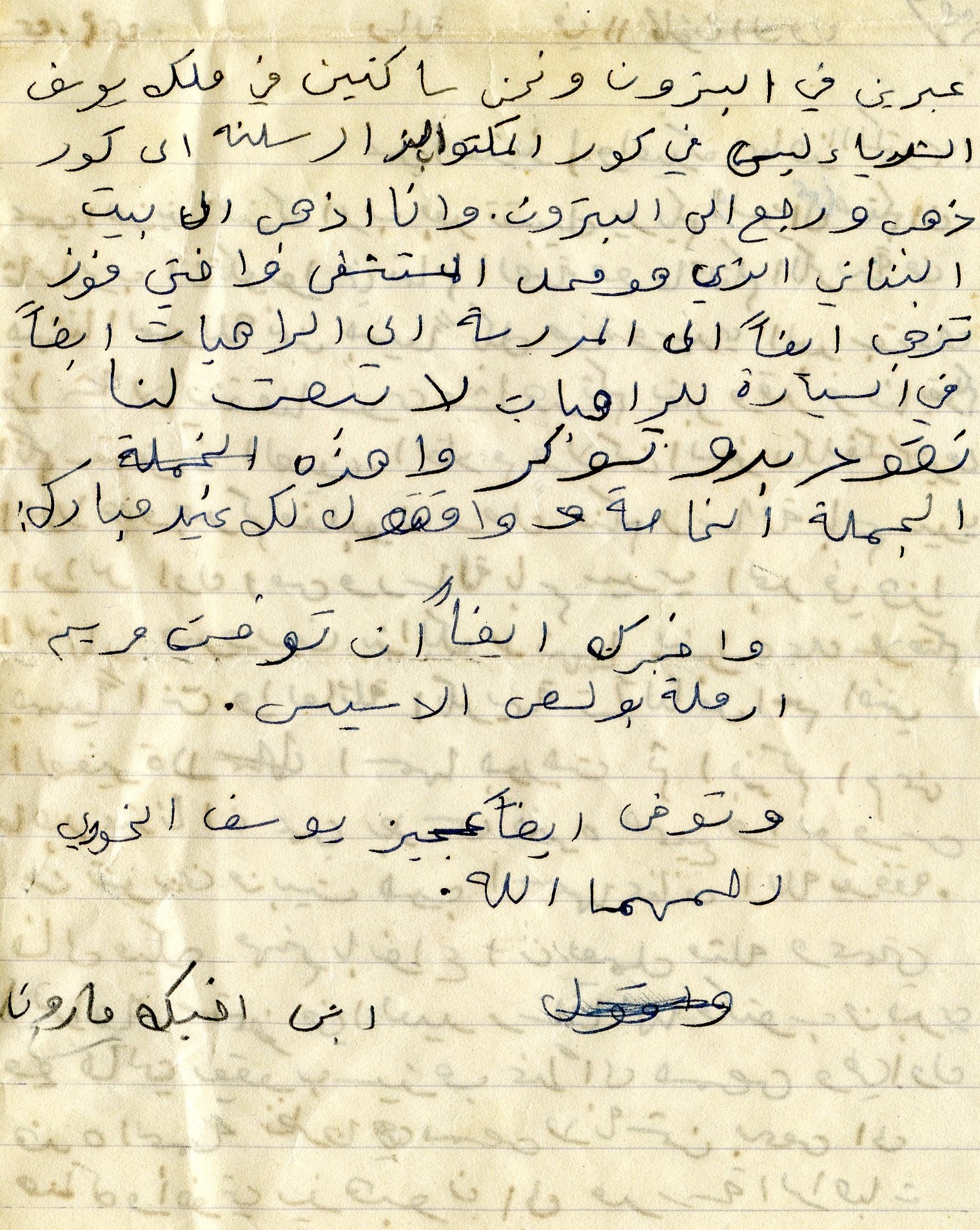}
    \caption{}
  \end{subfigure}  
\begin{subfigure}[t]{0.195\textwidth}
\includegraphics[width=\linewidth, height=3cm]{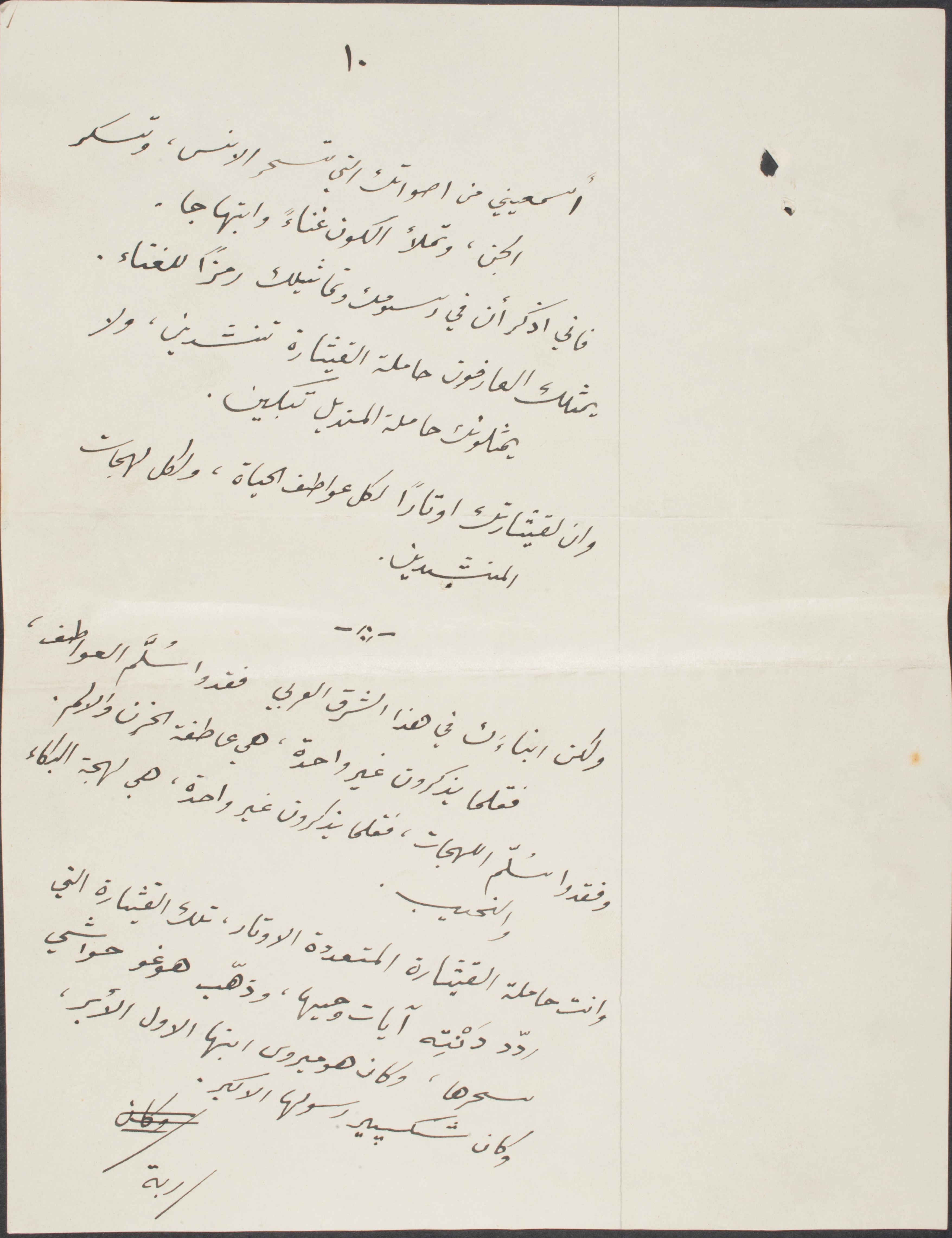}
    \caption{}
  \end{subfigure}    
\begin{subfigure}[t]{0.195\textwidth}
\includegraphics[width=\linewidth, height=3cm]{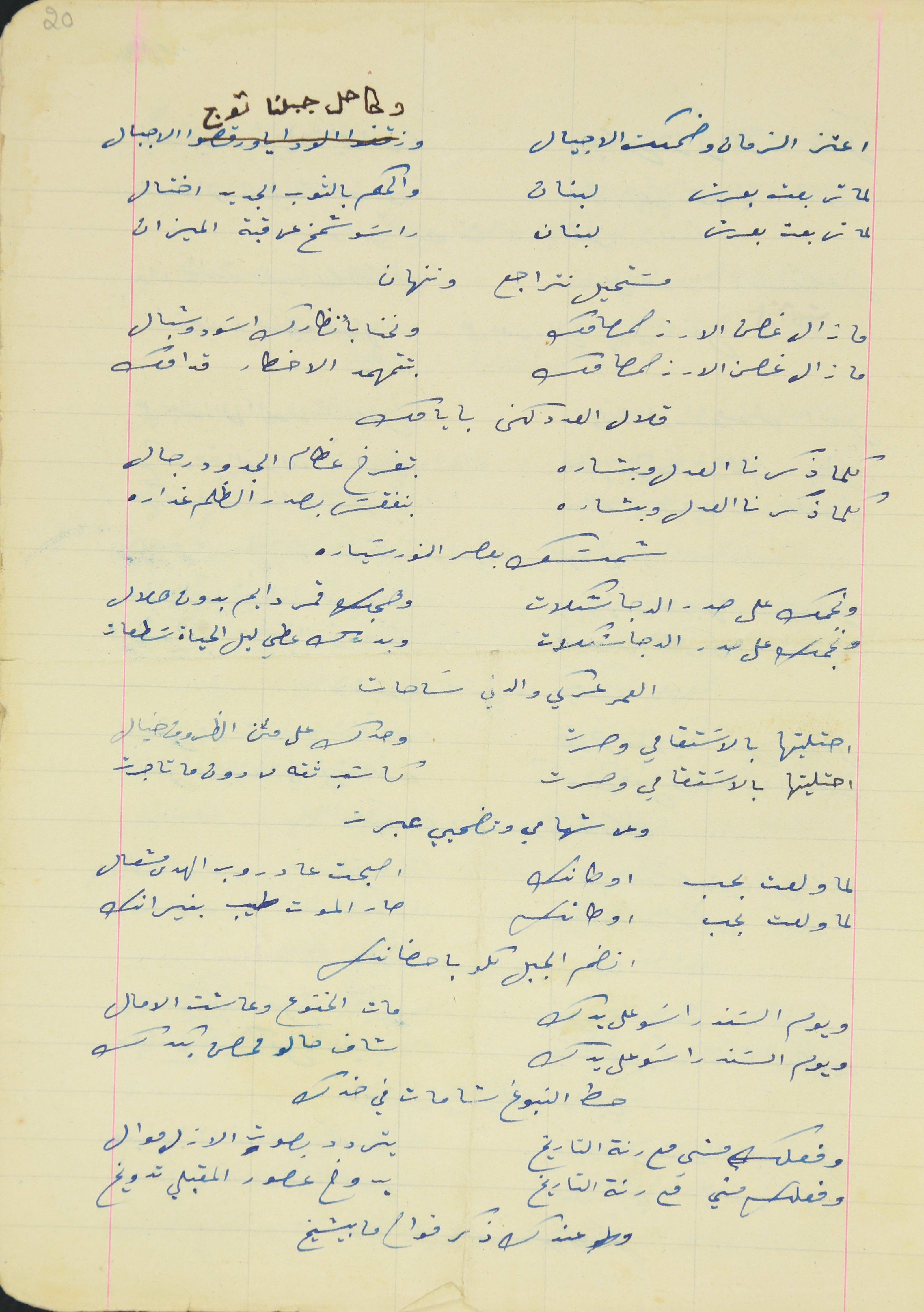}
    \caption{}
  \end{subfigure}  
\begin{subfigure}[t]{0.195\textwidth}   \includegraphics[width=\linewidth, height=3cm]{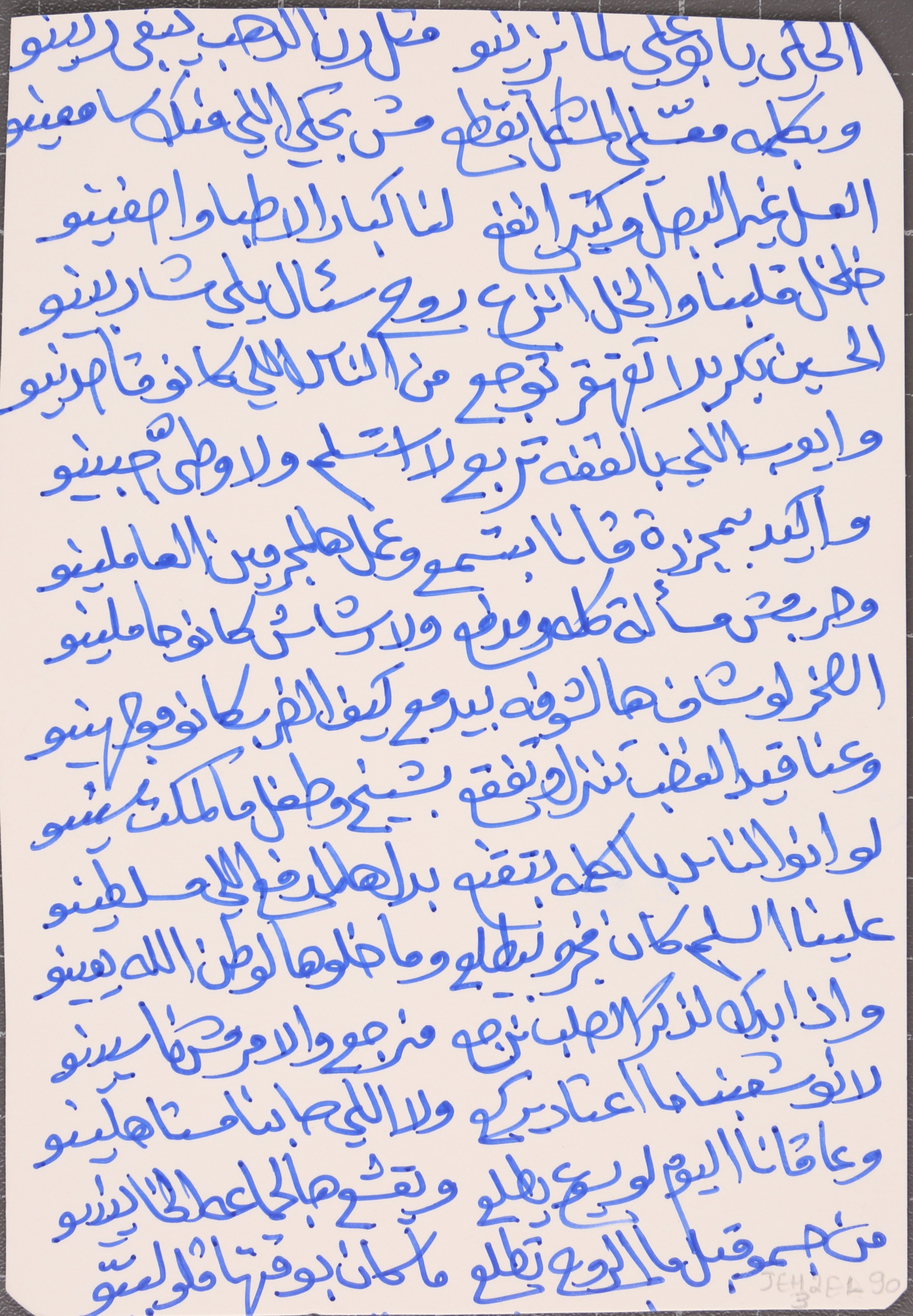}
    \caption{}
  \end{subfigure}    
\caption {Sample images of the Muharaf dataset from (a)~USEK Al Batroun Collection, (b)~KCLDS El-Khouri Letters Collection, (c)~KCLDS Ameen Rihani Collection, (d)~USEK Hanna Moussa Collection, and (e)~USEK Joseph El Hachem Collection.}
\label{fig:sample_images}
\end{figure*}

The Muharaf dataset consists of a diverse set of images, ranging from individual personal letters, poems, and dialogues to legal consensus records, correspondences, and church records. The manuscripts date from the early 19th century to the early 21st century. \changecolor{There are a total of \num{1644} images, \num{36311} text lines, and \num{4867} text regions including main text regions, headings, and floating text regions. The quality of page images varies, from writing on a clean white background to illegible sentences on creased pages with ink bleeds. A major part of the dataset with \num{1216} images, will be made publicly available and the remaining 428 images will be distributed under a proprietary license with permission from the owner.} A few samples from the public portion of this dataset are shown in Figure~\ref{fig:sample_images}. We used OpenAI's GPT APIs~\cite{openai_chatgpt_api} to generate English summaries and keywords corresponding to each manuscript page, which we provide for the general interest of the research community. 

In this paper, we also present the results of training a convolutional neural network~(CNN) based system of start, follow, and read (SFR) networks~\cite{sfr_2018}. Each network can easily fit in an 8~GB graphical processing unit~(GPU) card, making it ideal for deployment in a low-resource setting. \changecolor{Using a similar setup, an HTR system for languages based on the Arabic script like Urdu, Farsi, and Pashto can also be developed. Such systems can be initially trained using the Muharaf dataset and subsequently adapted and fine-tuned for the respective languages.} 

The rest of this paper is organized as follows. \magentaText{Section~\ref{section:arabic} outlines the relevant characteristics of the Arabic script and} Section~\ref{section:resources} overviews the existing Arabic datasets for OCR/HTR. Section~\ref{section:pipeline} describes the pipeline for collecting data and Section~\ref{section:corpus} details the dataset features, formats, and statistics. Section~\ref{section:sfr} provides the baseline results from our preliminary experiments on HTR. Section~\ref{section:conclusions} concludes the paper and discusses the limitations of this work along with future directions. 

\section{\magentaText{Characteristics of Arabic script in Muharaf}}\label{section:arabic}
\magentaText{The Arabic alphabet is believed to have its roots in the Nabatean alphabet from the third century~\cite{gruendler:1993}. The study of the origins of Arabic script and its evolution from classical Arabic to modern standard Arabic is a huge undertaking and still a subject of open scientific study and debate~\cite{rasam_2021}. For example, the early Arabic writings did not contain dots (ijam) but later the script evolved to include them. The Muharaf dataset has writing samples from the early 19th century to the early 21st century, a period during which the use of dots was well established. Most of Muharaf's samples are in the Arabic script Ruq`ah. Ruq`ah became the most common writing style in the late Ottoman period and early post-Ottoman period in the area we know today as the Middle East (which includes Egypt, but not North Africa).}

\magentaText{A few challenges related to OCR/HTR of the Arabic script include the following:\begin{enumerate}\item The shape of each character of the Arabic script depends upon its contextual position within a word. Many characters have four different shapes depending upon whether they are in their isolated, initial (at the beginning of a word), medial (in the middle of a word), or final forms (at the end of a word). This poses a challenge for the OCR/HTR system that has to recognize all the different forms of the same character. Table \ref{table:arabicForms} shows an example of 4 Arabic characters and their different forms.
\item Additional symbols in Arabic script include ijam, which are dots present above or below a character. Two characters may have the same basic shape but different numbers of dots to tell them apart. For example, the ijam or dots distinguish the bā' (one dot below) from tā' (two dots above), as shown in Table \ref{table:arabicForms}. An HTR system may misclassify a character because of the ijam.
\item Arabic language also has diacritics, which include ijam and tashkil. Tashkils are also called harakat. Harakat are short vowel marks in Arabic and are used to indicate the pronunciation of words. They are optional symbols and may not be present in the script. For example, the phrase without diacritics:  \RL{مجموعة بيانات محرف رائعة} turns into \RL{مَجْمُوعَةُ بَيَانَاتِ مُحَرَّفٍ رَائِعَةٌ} with diacritcs. The diacritics in the later phrase are the accent marks above or below the characters. In the context of an HTR system, the diacritics can get mislabeled for ijam and vice versa. They also increase the size of the character set that the system has to deal with.\end{enumerate}}

\begin{table}[t]
    \centering
    \caption{\magentaText{Isolated, initial, medial, and final forms of 4 different letters from the Arabic alphabet.}}    
    \label{table:arabicForms}
    
    \newcolumntype{L}[1]{>{\raggedright\arraybackslash}m{#1}}
    \newcolumntype{C}[1]{>{\centering\arraybackslash}m{#1}}
    \newcolumntype{R}[1]{>{\raggedleft\arraybackslash}m{#1}}

    
    \begin{filecontents*}{csv/arabicLetters.csv}
Letter|Isolated form|Initial Position|Medial Position|Final Position
bā'|\RL{ب}|\RL{بـ}|\RL{ـبـ}|\RL{ـب}
tā'|\RL{ت}|\RL{تـ}|\RL{ـتـ}|\RL{ـت}
lām|\RL{ل}|\RL{لـ}|\RL{ـلـ}|\RL{ـل}
mīm|\RL{م}|\RL{مـ}|\RL{ـمـ}|\RL{ـم}
\end{filecontents*}
\csvreader[separator=pipe,
        tabular={C{1.7cm} C{1.3cm} C{1.3cm} C{1.3cm} C{1.3cm}}, 
        table head=\toprule  Letter & Isolated Form&Initial Position& Medial Position&Final Position\\\midrule, 
        head to column names, 
      late after last line=\\\bottomrule,
    ]{csv/arabicLetters.csv}{} {
     \csvcoli & \csvcolii & \csvcoliii & \csvcoliv & \csvcolv}

\end{table}

\section{Publicly available Arabic resources}\label{section:resources}
\changecolor{The number of publicly available Arabic datasets is \magentaText{less} than those available for languages written in the Latin script.} Many Arabic datasets were tailored for specialized tasks, e.g., BADAM for baseline detection~\cite{badam_2019}, HADARA80P for word spotting~\cite{hadara_2014}, AHDB for detecting and recognizing numbers on legal checks~\cite{ahdb_2002}, and WAHD for writer identification~\cite{wahd_2017}. In our subsequent discussion, we focus on offline, handwritten datasets for Arabic HTR, where ground truths for text are available. We may classify offline handwritten HTR datasets into two categories:
\paragraph {Category 1 (scribed)} In this category, handwritten samples are obtained under controlled conditions by requesting scribes to copy paragraphs or lines of text provided to them. Such a scenario not only offers the flexibility of choosing text and frequency of words to transcribe, but also allows a researcher to choose writing styles, quality/texture of the paper, the writing implement such as pen or pencil, and, scanning/lighting conditions. Moreover, the ground truth is predetermined with the caveat that the handwritten pages have to be manually verified to ensure that a writer has not made any mistakes. The widely used IAM dataset of handwritten English sentences~\cite{iam_2002} falls into this category. The same goes for the French {RIMES} dataset~\cite{rimes_2008} and its latest update~\cite{rimes_2024}. Such datasets are better suited for the HTR of contemporary documents. 

The earlier Arabic handwritten OCR datasets in this category consisted of word images and their corresponding transcriptions. Examples include the IFN/ENIT dataset~\cite{ifn_enit_2002}, which has \num{26549} images, compiled from a vocabulary of \num{937} Tunisian town names. Similarly, the IFN/Farsi~\cite{ifn_farsi_2008} has \num{7271} word images of the names of Iranian cities and provinces. 

To the best of our knowledge, the KHATT dataset~\cite{khatt_2012} is the first Arabic dataset with paragraph-level handwriting and corresponding ground truth. \num{1000} writers filled out forms by copying preprinted text on the same page. Paragraphs were later segmented automatically and manually verified. The MADCAT Phase 1--3 training sets~\cite{madcat_2012} consist of an overall \num{42047} scanned handwritten page images. Writers were asked to copy documents by hand using various writing styles (fast, normal, and careful), on lined or unlined paper using a pen or pencil. However, this dataset is not freely available to the public, which restricts its use. 

\paragraph {Category 2 (original)} This category comprises scanned genuine handwritten documents that have been annotated and transcribed by individuals fluent in a respective language. \changecolor{For historical manuscripts, the expertise of historians or linguists may be required.} The well-known ICDAR 2017 HTR competition dataset~\cite{icdar_2017} of early modern German language from the READ project belongs to this group with page-level transcriptions (instead of line-level) of more than \num{10000} images. Our proposed Muharaf dataset also falls into this category, comprising a collection of historic manuscripts that primarily range from the late 19th to the mid-20th century.  

\changecolor{RASM~\cite{rasm_2018} and RASAM~\cite{rasam_2021} are two Arabic datasets of scanned original historic manuscripts. Both datasets have annotated text regions and text lines along with their corresponding transcriptions.} A more recent dataset is the Historic Arabic HTR dataset~\cite{htr_dataset_2024} with a collection of 40 pages and their corresponding page-level ground truths. Table~\ref{table:datasets} summarizes key Arabic HTR datasets that we are aware of.

\begin{table}[tp]
    \centering
    \caption{\magentaText{An overview of key Arabic HTR datasets.}} 
    \label{table:datasets}
    
    \newcolumntype{L}[1]{>{\raggedright\arraybackslash}m{#1}}
    \newcolumntype{C}[1]{>{\centering\arraybackslash}m{#1}}
    \newcolumntype{R}[1]{>{\raggedleft\arraybackslash}m{#1}}

    
    \begin{filecontents*}{csv/arabicOCRDatasetsRevised.csv}
Dataset|Category|Annotated Text Lines|Total Writers| Vocabulary/Composition
IFN/ENIT (2002) \cite{ifn_enit_2002}|1|\xmark|\num{411}|\num{937} Tunisian town names. \num{26549} word images.
AHDB (2002)~\cite{ahdb_2002}|1|\xmark|100|Numbers and phrases used to express numbers on legal checks, \num{20} most frequently used Arabic words, free hand paragraphs.
IFN/FARSI (2008) \cite{ifn_farsi_2008}|1|\xmark|\num{600}| \num{1080} Iranian city/province names. \num{7271} word images.
KHATT (2012) \cite{khatt_2012}|1|\cmark|\num{1000}|\num{2000} unique text + \num{2000} similar text paragraphs. \num{1000} writers. Paragraphs segmented automatically and text corrected manually.
AHTID/MW (2012) \cite{ahtid_mw_2012}|1|\cmark|\num{53}|Open vocabulary. \num{3710} annotated lines. \num{22896} words.
MADCAT (2012)~\cite{madcat_2012} |1|\cmark|311$^a$|Document source: weblogs, newswires, and newsgroups. \num{42047} page images written by scribes under controlled conditions.
HADARA80P (2014)~\cite{hadara_2014}|2|\xmark|1$^b$|\num{80} pages scanned from the Taaun book (1430 AD). Annotation of pages, text blocks, and words. Ground truth for \num{16720} individual words.
VML-HD (2017)~\cite{vml_hd_2017}|2|\xmark|-|\num{680} pages from historic manuscripts with dates ranging 1088--1451. \num{159149} word annotations.
RASM (2018)~\cite{rasm_2018}|2|\cmark|-|\num{120} historic scientific manuscript pages. \num{2613} annotated text lines. Text regions segmented. 
RASAM (2021)~\cite{rasam_2021}|2|\cmark|-|\num{300} historic Maghrebi script manuscript pages from 10th century. \num{7540} annotated text lines. Text regions segmented.
Historical Arabic HTR (2024) \cite{htr_dataset_2024}|2|\xmark|-|\num{40} pages from 8 different historical books. Transcriptions at the page level.
Muharaf (this paper)|2|\cmark|-|\num{1644} (\num{1216} public, \num{428} restricted) pages of historic manuscripts from 1800--2018. Text regions segmented. \num{36311} (\num{24495} public, \num{11816} restricted) annotated text lines.
\end{filecontents*}
\csvreader[separator=pipe,
        tabular={C{1.7cm} C{1.1cm} C{1.1cm} L{1cm} L{6.9cm}}, 
        table head=\toprule  Dataset & Category &Annotated Text Lines& Total Writers & Vocabulary/Composition\\\midrule, 
        head to column names, 
        late after line=\\\addlinespace,
      late after last line=\\\bottomrule,
    ]{csv/arabicOCRDatasetsRevised.csv}{} {
    \ifnum\thecsvrow=12 \midrule\fi  \csvcoli & \csvcolii & \csvcoliii & \csvcoliv &\csvcolv}

\magentaText{\footnotesize{$^a$ From unique subject IDs in scribe\_demographic file.\\$^b$ Based on text-independent features for handwriting analysis~\cite{hadara_2013}.\\- Total number of writers is unspecified or unknown due to the nature of the dataset.}}
\end{table}

Table~\ref{table:line_summaries} shows the line-level statistics of the IAM English dataset and other publicly available Arabic datasets containing line-level text annotations. Out of the Arabic datasets that are publicly accessible, it is evident that Muharaf contains the largest number of annotated text lines. 

\begin{table}[tp]
    \centering
    \caption{A comparison of various HTR datasets \magentaText{in terms of total pages, text regions, and total lines}.}
    \label{table:line_summaries}
    \newcolumntype{L}[1]{>{\raggedright\arraybackslash}p{#1}}
    \newcolumntype{C}[1]{>{\centering\arraybackslash}p{#1}}
    \newcolumntype{R}[1]{>{\raggedleft\arraybackslash}p{#1}}

    
\begin{filecontents*}{csv/line_summaries.csv}
Dataset|Page Count|Text Regions/ Paragraphs|Line Count
IAM \cite{iam_2002}|\num{1539}| \num{1539}|\num{13353}
RASAM \cite{rasam_2021}|\num{300}|\num{676}|\num{7540}
RASM \cite{rasm_2018}|\num{120}|\num{132}|\num{2613}
KHATT \cite{khatt_2012}|\num{4000}$^a$|\num{4000}$^a$|\num{13435}
Muharaf-public |\num{1216}|\num{3479}$^b$|\num{24495}
Muharaf-restricted |\num{428}|\num{1388}$^b$|\num{11816}
Muharaf |\num{1644}|\num{4867}$^b$|\num{36311}
\end{filecontents*}
\csvreader[separator=pipe,
        tabular={ C{3cm}  C{2cm}  C{2.3cm}  C{2cm} }, 
        table head=\toprule Dataset & Page Count &Text Regions& Line Count \\\midrule, 
        head to column names, 
        late after line=\\, 
        late after last line=\\\bottomrule,
    ]{csv/line_summaries.csv}{} 
    {
    \ifnum\thecsvrow=5\midrule\fi  \csvcoli & \csvcolii & \csvcoliii & \csvcoliv }
    
    \footnotesize{$^a$ Includes fixed and unique text paragraphs.
    \\$^b$ Includes main text regions, headings, and floating text regions.}
\end{table}

\paragraph{\magentaText{Contrasting features of Muharaf and other available Arabic datasets}}
\magentaText{The existing Arabic datasets are a valuable resource for the research community. Muharaf supplements them with its own unique features and characteristics, discussed below:
\begin{enumerate}
\item As shown in Table \ref{table:line_summaries}, Muharaf has the largest number of line images as compared to IAM, RASAM, RASM, and KHATT.
\item KHATT and MADCAT datasets are category 1 (scribed) datasets, where writers were given text to write under controlled experimental conditions. Muharaf is a category 2 (original) dataset, where original historic manuscripts were scanned and transcribed.
\item RASAM and RASM are two Arabic datasets of scanned original historic manuscripts from category 2. The main difference between these two datasets and Muharaf lies in the Arabic script with which they were written (see points 4 and 5 below). Another difference is that RASAM and RASM handwritten pages belong to books, with calligraphic handwriting that is very neat and uniform across pages, written in straight horizontal lines. They were written by scholars in their respective fields. Muharaf includes informal/personal styles of writing, which were very common in the 19th and 20th centuries. The samples vary from very neat to barely legible writing. The handwriting samples of the same individual can be different over different documents or letters. Moreover, the text lines can be slanted upward or downward instead of horizontally straight lines.
\item RASAM has three types of manuscripts from the 10th century. They are scanned pages of books, which were written in the ``Meghrebi script'' also known as the ``Round script'' \cite{rasam_2021}. As the name suggests, this script has very rounded shaped letters. In contrast, Muharaf's documents are mostly Ruq`ah script, which is used for everyday or casual writing. It is composed of straight, short lines, and simple curves.
\item RASM has 4 different types of manuscripts of scientific writings from the 8th century to the 19th century. Muharaf has images from 50 different collections, each collection having one or more writers. Like RASAM, RASM's handwriting styles are calligraphic, very neat, and uniform across all pages as opposed to Muharaf, where the writer may not have very careful or readable handwriting.
\end{enumerate}
The OCR of text from different Arabic datasets like RASAM and RASM has its own challenges and is by no means an easy feat. We intend to supplement the existing Arabic datasets with a variety of handwritten images with the goal of digitizing handwritten documents from the late 19th century to the mid-20th century. 
}

\section{Data collection process}\label{section:pipeline}
\begin{figure}[tbp]
\centering
\includegraphics[width=1\textwidth]{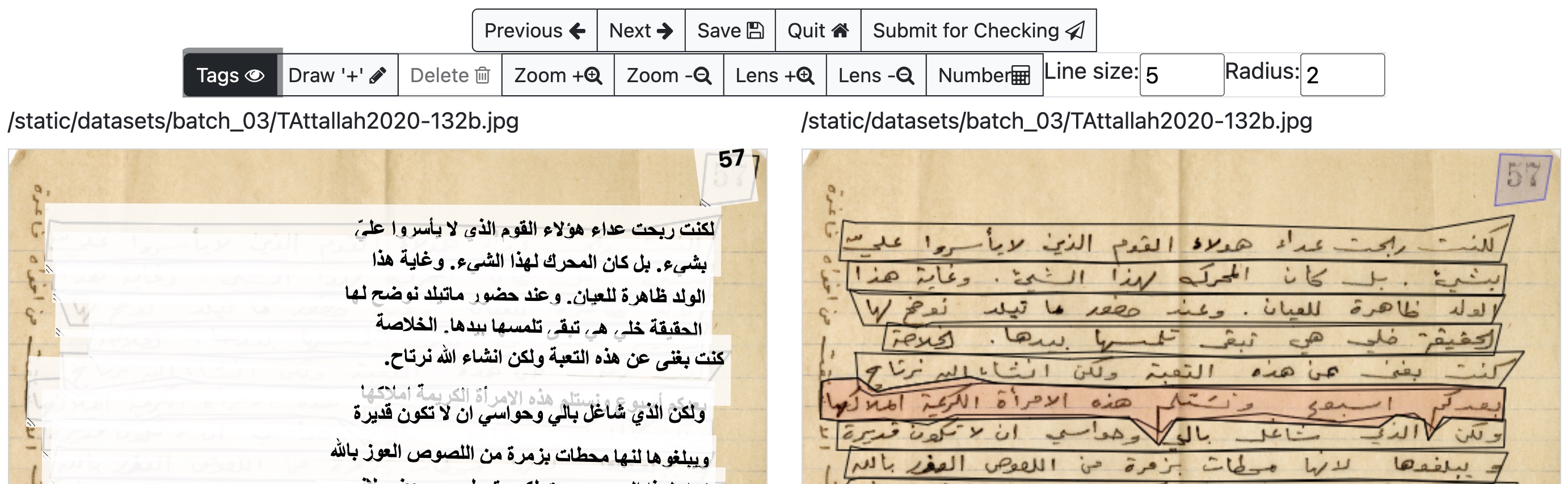}
\caption {\changecolor{A screenshot showing the graphical user interface of the ScribeArabic software used for labeling page images.}}
\label{fig:tool_screen_shot}
\end{figure}

We developed an image-labeling software named ScribeArabic\footnote{For code and a link to demo, please see \url{https://github.com/MehreenMehreen/ScribeArabic}} specialized for annotating and transcribing Arabic page images of the Muharaf dataset. This software allows a user to annotate text lines in a browser window and transcribe them in a panel next to it. A separate module has the option to label various page elements. A screenshot of this software is shown in Figure~\ref{fig:tool_screen_shot}. \magentaText{We have made the source code for ScribeArabic publicly available (see Section~\ref{appendix:source_code} of the supplementary material for all repository links).}

The following steps are involved in labeling a page image using ScribeArabic:
\begin{enumerate}
\item Line annotation: Marking a polygonal boundary around each text line.
\item Line transcription: Entering the ground truth transcription for each annotated text line.
\item Defining page elements (if needed): Marking, labeling, and tagging basic page elements such as headings, page numbers, floating text, and graphics.
\item Quality assurance (QA): Verifying that the labeling of a page image is correct.
\end{enumerate}

Transcribing historic handwritten Arabic manuscripts primarily requires experts in archival Arabic, though Steps 1 and 3 permit less specialized involvement. Step 1 for line annotations can be performed by non-Arabic speakers with basic knowledge of the Arabic script. Step 3 for tagging basic page elements generally does not require any Arabic knowledge. However, only an Arabic expert can do the transcriptions and the QA steps. 

We assembled a team of expert Arabic speakers to transcribe historic Arabic manuscripts with technical support from machine learning researchers. The text lines of the first 180 images of our curated dataset were annotated by non-Arabic speakers using the ScribeArabic software. A Lebanese Arabic history professor then manually entered the transcriptions into an Excel sheet. We chose Excel for entering transcriptions because initially ScribeArabic supported only line annotations. For the next \num{1400}+ images, we used our upgraded ScribeArabic software to allow the direct input of transcriptions in a browser window. The annotations and transcriptions for these \num{1400}+ images were performed by two native Lebanese Arabic speakers adept at archival Arabic. Their transcriptions were checked by a third Lebanese Arabic expert who was also a historian.
\magentaText{Section~\ref{paragraph:qualifications} on page~\pageref{paragraph:qualifications} has more details of the qualifications of the annotation and transcription team.}

Besides the ScribeArabic software, we employed deep learning to speed up the data collection process. After labeling the first 500+ page images, we trained the SFR system to do a full-page HTR. For transcribing subsequent images, we provided the preliminary line annotation and transcription results from SFR to the transcribers for manual correction. This streamlined into an iterative process of training SFR with more data and manually correcting the line annotation and transcription on unseen images. The HTR system showed considerable performance improvement over time as we acquired more data. Section~\ref{section:sfr} has more details of these experiments. 

\section {Muharaf file formats and features}\label{section:corpus}

The ground truths for the Muharaf dataset are provided using the page analysis and ground-truth elements (PAGE) XML format~\cite{url-page} and our own Javascript object notation (JSON) file format. \changecolor{The PAGE-XML format is supported for compatibility with Pattern Recognition and Image Analysis~(PRImA) Research Lab's Aletheia tool~\cite{aletheia_2011} that allows users to open, view, and edit the annotations, transcriptions, and page elements of a document image.} They can also view the images using the PRImA Research Lab's PAGE-viewer~\cite{url-page-viewer}. We briefly explain the file formats next.

\subsection {PAGE-XML format for OCR/HTR datasets}\label{section:xml}
\begin{figure}[t]
\centering
\includegraphics[width=0.5\textwidth]{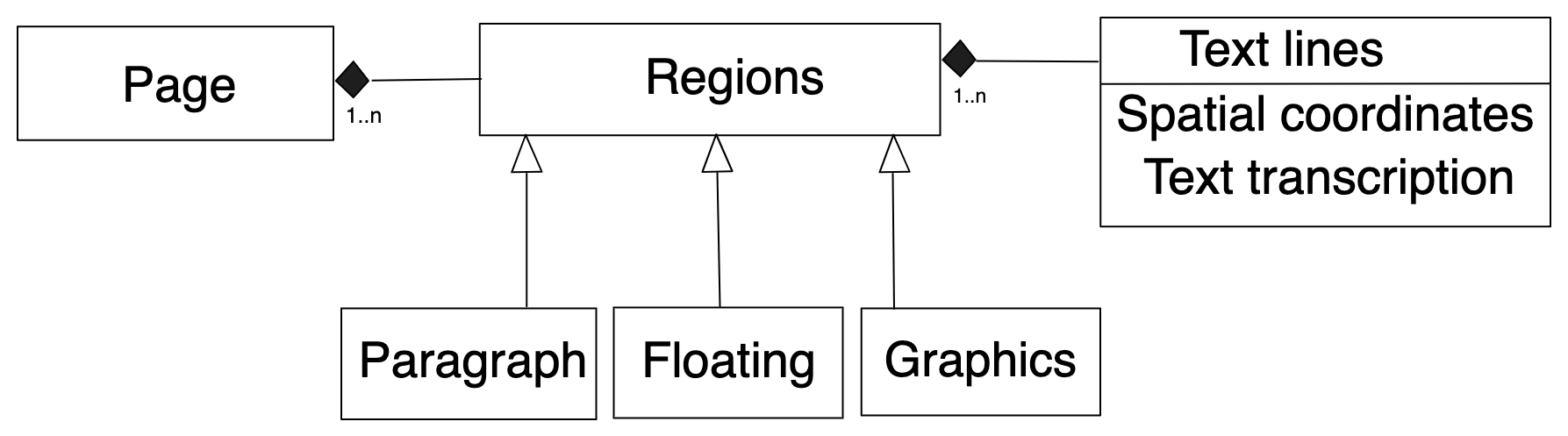}
\caption {A class diagram of the hierarchy defined by PAGE-XML format. The black diamond indicates the composition relation between two classes. The arrows indicate inheritance. Note: We are using only a subset of the PAGE schema for our dataset.}
\label{fig:class_diagram}
\end{figure}

The PAGE-XML format is an XML-based page image representation framework introduced by the PRImA Research Lab~\cite{page-2010}. This format incorporates various image characteristics as well as the information on page layout and its contents at various levels of granularity. The official, full description of the XML schema is hosted at \url{https://www.primaresearch.org}. A conceptual class diagram of the subset of the PAGE-XML hierarchy used in the Muharaf dataset is shown in Figure~\ref{fig:class_diagram}. Where applicable, the following labeled regions are present in the PAGE-XML file of each image \magentaText{(see Section~\ref{appendix:pageElements} of the supplementary material for illustrative examples).}
\begin{itemize}
\item Paragraph regions: Main body of text on the page.
\item Floating regions: Regions of text outside the normal flow of text. Examples include footer and margin texts.
\item Graphics regions: These may include stamps, letterheads, and logos. \changecolor{PAGE-XML format allows these regions to contain text lines.}
\item Page number regions: Regions containing page numbers.
\item Signature regions: Page areas containing names and signatures.
\end{itemize}

\changecolor{While a large majority of lines in our dataset are handwritten cursive, there are occasional printed characters found on letterheads, logos, or stamps. We have tagged any instance of this printed text as either printed-regular, printed-bold, or printed-italics. We also made the following annotation rules:}
\begin{itemize} 
\item Most of the crossed-out text is annotated, although its transcription is not included.
\item We did not manually mark the baseline of the text, but algorithmically computed a mid-line passing through the polygon enclosing the text line. This appears in the XML file under the <UserDefined> tag. This line can be used to determine the orientation of the line and reading direction, which is right-to-left for Arabic and left-to-right for English.
\end{itemize}

\subsection {JSON file formats used by the ScribeArabic software}
The JSON format is a simple data-structure dictionary with key--value pairs used to represent entities. Our main data collection process involves the use of ScribeArabic that natively supports the JSON format. We provide two JSON files of different types for each image. The first type has only the line coordinates and their corresponding transcriptions in the ``coord'' and ``text'' keys, respectively. These files have the ``\_annotate.json'' suffix and are intended for researchers working only on text-line segmentation and line-level HTR/OCR.

\changecolor{For each image, we also provide a second type of JSON file with the ``\_tagged.json'' suffix. It contains the labeling of various page elements along with individual text-line annotations and transcriptions. Its format is described in detail in Section~\ref{appendix:json} of the supplementary material. In our source code, we provide a script to convert the second type of JSON files to the PAGE-XML format. We have also included a custom viewer that reads the JSON file and renders the various page elements of the corresponding image in different colors. Users can use the viewer to browse through images in a directory (Section~\ref{appendix:source_code} of the supplementary material has more details).
}

\subsection {Additional characteristics of the Muharaf dataset}

\changecolor{Each page image of the Muharaf dataset is part of a specific collection from either the archives of Phoenix Center for Lebanese Studies at USEK or KCLDS at NC State. In the digital archives, a librarian or archivist places all the images of document manuscripts from the same writer, period, or category in a collection. The filenames of all images from the same collection share a prefix that uniquely defines the collection. } 

\changecolor{A list of 50 collections in the Muharaf dataset and their associated characteristics is presented in Section~\ref{appendix:collections} of the supplementary material. This list includes an approximate period (if known) for a collection, the total number of image files in a collection, and the top four keywords that describe the collection. The keywords were generated by querying OpenAI's GPT 3.5 APIs~\cite{openai_chatgpt_api} with the corresponding Arabic text transcription. The steps and prompts are provided in Section~\ref{appendix:chatgpt} of the supplementary material.}   

\subsection {Line images and their text transcriptions}

As handwritten text lines can be slanted or curved, we used the line warping code of SFR~\cite{sfr_2018} to convert them to straight line images. The Muharaf dataset includes a separate directory with text-line images stored in the PNG format and their corresponding text transcriptions stored in the plain text format. This directory is enclosed for the convenience of researchers experimenting with line-level OCR/HTR. They can use this portion of the dataset without having to apply any processing for extracting them from the raw page images or warping them to a straight horizontal image grid. All line images have a height of \num{60} pixels and a width ranging from \num{60}--\num{2400}  pixels (with an average of \num{576} pixels). 

\begin{figure} 
\centering
\includegraphics[width=1\textwidth]{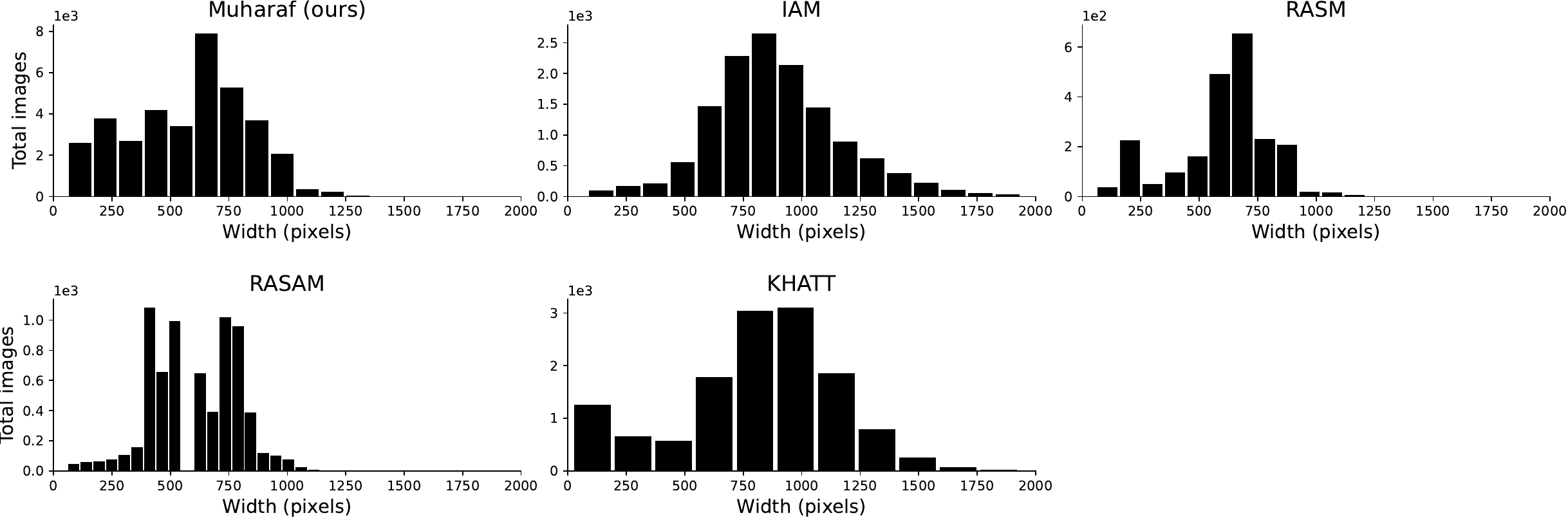}
\caption {The histograms of the widths of line images for various datasets. Line images were warped to a height of \num{60} pixels while maintaining the aspect ratio.}
\label{fig:width_dist}
\end{figure}

Figure~\ref{fig:width_dist} compares the distributions of image widths among Muharaf and other datasets. \changecolor{The plots reveal that the widths of line images in the IAM dataset are more normally distributed given that the sentences in the dataset were manually selected. The Muharaf, RASAM, and RASM datasets, in contrast, are based on actual documents and all of them have fewer lines that are more than \num{1000} pixels wide. Nevertheless, Muharaf has comparatively more line images of shorter widths.}

\section{Baseline HTR results and discussion}\label{section:sfr}
The SFR~\cite{sfr_2018} system can be trained to do a full-page HTR. It comprises three CNN-based networks: 
\begin{enumerate}
\item Start of line (SOL) network: A CNN for detecting the coordinates of the start of each text line of a page image.
\item Line follower (LF) network: A CNN that tracks and follows the trajectory of a text line. Starting from the SOL position, this network iteratively predicts the next coordinates and orientation of the text line given its previous coordinates and orientation within a small viewing window. 
\item Handwriting recognition (HW) network: A CNN--Bidirectional long short-term memory~(BLSTM) network trained using connectionist temporal classification~(CTC) loss~\cite{ctc_2006}.  
\end{enumerate}
\changecolor{Each network in the SFR system can be trained independently, even in a low-resource environment (described in Section~\ref{section:evolution}). We used the Muharaf dataset to train the SFR system for full page HTR using a split of \num{1500}, \num{50}, and \num{96} images for training, validation, and testing, respectively. The metrics for evaluation are the character error rate (CER) and word error rate (WER), both computed using Levenshtein distance~\cite{levenshtein:1966} normalized by the length of the string representing the ground truth. The experiments were repeated three times using a different random split of the training, validation, and test sets. Table~\ref{table:trial_15} shows the statistical results of the three experiments in terms of the sample mean and standard deviation. Table~\ref{table:trial_15} also reports the statistical results of training on the public part of the Muharaf dataset.} The line-level CER and WER reflect the performance of the HW network on individual pre-segmented text lines extracted using the spatial coordinates of the ground truth annotations. Both error rates are higher than the page-level CER and WER due to the presence of many lines with isolated numbers or single words. If the system makes a mistake on these lines, the CER/WER for these lines can jump to as high as 1.0, contributing to a higher average error rate.

\begin{table}[htbp]
    \centering
    \caption{HTR results repeated over three random splits of Muharaf. \magentaText{Both page level and line level results are included.}}   
    \label{table:trial_15}
    \newcolumntype{L}[1]{>{\raggedright\arraybackslash}p{#1}}
    \newcolumntype{C}[1]{>{\centering\arraybackslash}p{#1}}
    \newcolumntype{R}[1]{>{\raggedleft\arraybackslash}p{#1}}
    \begin{filecontents*}{csv/trial_15.csv}
Type|Split||CER|WER
Muharaf-public|$(1100, 50, 66)$|Page|\num{0.157(0.008)}|\num{0.398(0.007)}
||Line|\num{0.181(0.009)}|\num{0.430(0.011)}
Muharaf|$(1500, 50, 96)$|Page|\num{0.134(0.007)}|\num{0.353(0.012)}
||Line|\num{0.149 (0.004)}|\num{0.380(0.004)}
\end{filecontents*}
\csvreader[separator=pipe,
        tabular={ L{2cm} C{4cm} C{1.5cm}  C{2.2cm}  C{2.2cm} }, 
        table head=\toprule Dataset& Split&Level& CER & WER \\&{(Train, Validate, Test)}&&&\\\midrule, 
        head to column names, 
        late after line=\\, 
        table foot = \bottomrule,
    ]{csv/trial_15.csv}{} 
    {    \ifnum\thecsvrow=3\midrule\fi
    \multirow{2}{*}{\csvcoli} 
    & \multirow{2}{*}{\csvcolii} & \csvcoliii &\csvcoliv &\csvcolv 
    }
\end{table}

\subsection{Evolution of HTR performance as more data were collected}\label{section:evolution}
\begin{figure}[b]
\centering
\includegraphics[width=1\textwidth]{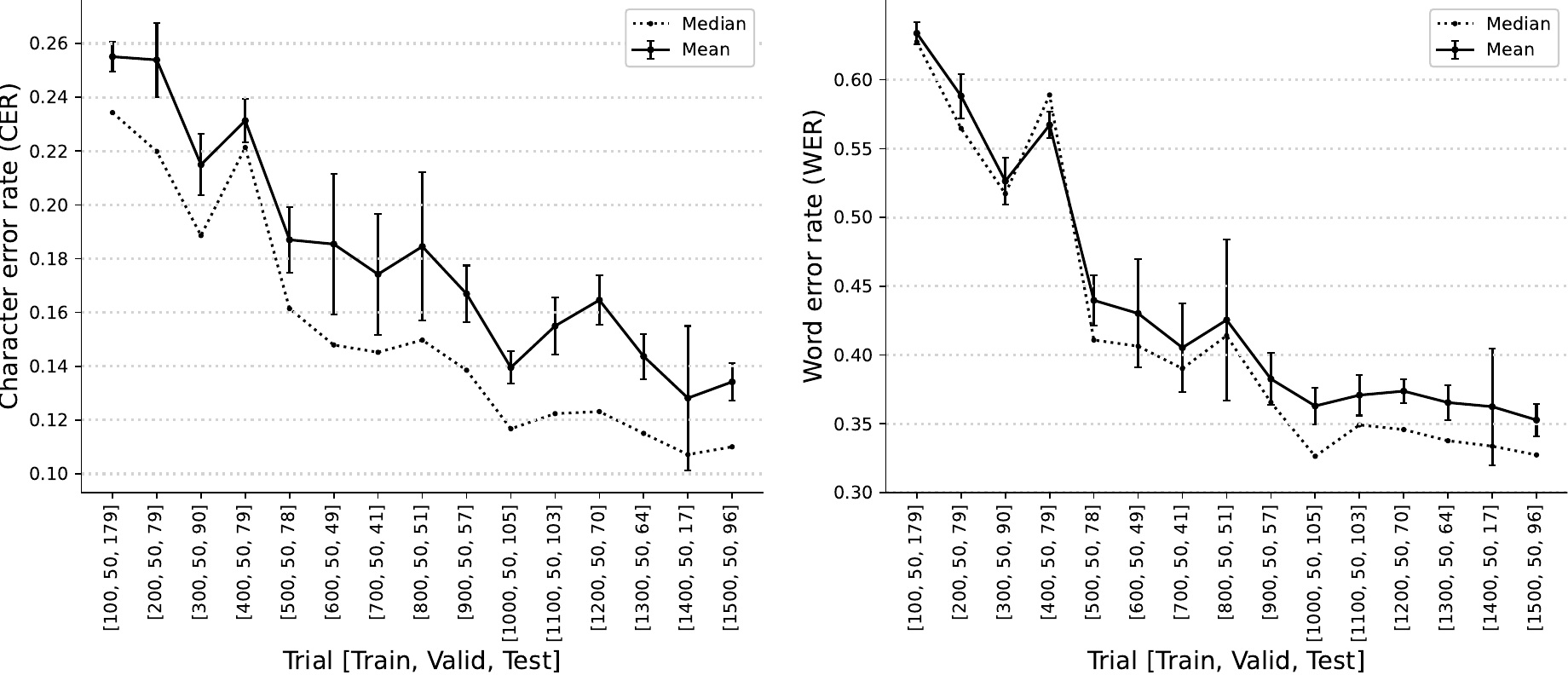}
\caption {The evolution of the page-level error rates for SFR as we collected more data for creating the Muharaf dataset. The horizontal axis shows the numbers of images in the train, validation, and test sets. The results are averaged over four different random splits for trials 1--8 and three different random splits for trials 9--15.}
\label{fig:trial_results}
\end{figure}

We continued to train SFR while the data collection process was going on. Figure~\ref{fig:trial_results} illustrates how the system evolved. The plots show the CER and WER over the course of 15 trials. Each trial was repeated three or four times and the sample mean and median statistics are reported. The train, validation, and test sets comprised images for which we had the ground truths available at that time, and hence, the total number of images in the test set of each trial varies. The CER plot reveals that after trial 5 (with 500 training images) the error rate dropped below 20\%. \changecolor{We ran the first 11 trials either on Ubuntu 22.04.4 LTS with two Nvidia cards (RTX 3060 12\,GB and RTX 2080 8\,GB) or on Ubuntu 20.04.6 LTS with a single RTX 4090 24\,GB card. Trials 12--14 were run on the NC State's high-performance computing~(HPC) cluster.}


\section{\magentaText{Conclusion and Future Work}} \label{section:conclusions}

In this paper, we have introduced a new machine learning dataset of historic handwritten Arabic manuscript images with their annotations and transcriptions. This dataset is a rich collection of a wide range of images with different characteristics. It includes diverse writing styles on a variety of paper backgrounds, ranging from clear handwriting to instances of torn paper or ink bleeds. There are personal notes, diary pages, legal correspondences, financial records, church records, etc. within this dataset, each telling stories of the past and offering a wealth of valuable information. The Muharaf dataset can be used to train a wide variety of systems such as HTR, text-line segmentation, layout detection, and writer identification. 

\magentaText{
To the best of our knowledge, Muharaf is the largest publicly available Arabic dataset comprising fully annotated and transcribed historical manuscript pages at the text-line level. However, this collection is not devoid of limitations. The process of identifying all the writers and the exact timeline for each document will continue after the initial release of the dataset. For documents where the writer's information cannot be extracted, e.g., the case where a scribe penned a legal document or a church record, one may want to define categories of different writers and writing styles. 
Another area of interest is the use of Muharaf transcriptions for the extraction of linguistic knowledge and the identification of the colloquial form of the Arabic language used in a particular period. Language models based on this information may improve the performance of the HTR system. We invite researchers in the document analysis and OCR community to utilize this dataset and advance the state of the art.}

\begin{ack}
We thank Stephen Randall Filios from Family Search for initiating discussions and providing feedback on tagging page elements in document images. \blueText{We thank Elham Abdallah who is the Assistant University Librarian at USEK for providing support and coordinating work between NC State and USEK.}

\magentaText{We acknowledge the computing resources provided by North Carolina State University High-Performance Computing Services Core Facility (RRID:SCR\_022168). We also thank Andrew Petersen for his assistance and technical guidance on running jobs on the HPC.}

This work was supported in part by the National Endowment for the Humanities (FAIN: ZPA-283823-22), Family Search, and the ECE Undergraduate Research Program at NC State.
\end{ack}

\bibliographystyle{plain} 
\bibliography{neurips}

\begin{thebibliography}{10}

\bibitem{khaleej_2005}
Mourad Abbas and Kamel Sma{\"i}li.
\newblock Comparison of topic identification methods for {A}rabic language.
\newblock In {\em Proceedings of the Recent Advances in Natural Language Processing}, pages 14--17, 09 2005.

\bibitem{watan_2011}
Mourad Abbas, Kamel Smaïli, and Daoud Berkani.
\newblock Evaluation of topic identification methods on {A}rabic corpora.
\newblock {\em Journal of Digital Information Management}, 9:185--192, 10 2011.

\bibitem{wahd_2017}
Alaa Abdelhaleem, Ahmed Droby, Abedelkader Asi, Majeed Kassis, Reem~Al Asam, and Jihad El-sanaa.
\newblock {WAHD}: A database for writer identification of {A}rabic historical documents.
\newblock In {\em Proceedings of the 1st International Workshop on {A}rabic Script Analysis and Recognition ({ASAR})}, pages 64--68, 2017.

\bibitem{ahdb_2002}
Somaya Al-Ma'adeed, Dave Elliman, and Collins~A. Higgins.
\newblock A data base for {A}rabic handwritten text recognition research.
\newblock In {\em Proceedings of the Eighth International Workshop on Frontiers in Handwriting Recognition}, pages 485--489, 2002.

\bibitem{metahtr_2021}
Ayan~Kumar Bhunia, Shuvozit Ghose, Amandeep Kumar, Pinaki~Nath Chowdhury, Aneeshan Sain, and Yi-Zhe Song.
\newblock {MetaHTR}: Towards writer-adaptive handwritten text recognition.
\newblock In {\em Proceedings of the {IEEE/CVF} Conference on Computer Vision and Pattern Recognition {(CVPR)}}, pages 15825--15834, 2021.

\bibitem{rasm_2018}
Christian Clausner, Apostolos Antonacopoulos, Nora Mcgregor, and Daniel Wilson-Nunn.
\newblock {ICFHR} 2018 competition on recognition of historical {A}rabic scientific manuscripts --- {RASM2018}.
\newblock In {\em Proceedings of the 16th International Conference on Frontiers in Handwriting Recognition {(ICFHR)}}, pages 471--476, 2018.

\bibitem{aletheia_2011}
Christian Clausner, Stefan Pletschacher, and Apostolos Antonacopoulos.
\newblock Aletheia --- {A}n advanced document layout and text ground-truthing system for production environments.
\newblock In {\em Proceedings of the International Conference on Document Analysis and Recognition ({ICDAR})}, pages 48--52, 2011.

\bibitem{datasheet_2021}
Timnit Gebru, Jamie Morgenstern, Briana Vecchione, Jennifer~Wortman Vaughan, Hanna Wallach, Hal~Daum\'{e} III, and Kate Crawford.
\newblock Datasheets for datasets.
\newblock {\em Commun. ACM}, 64(12):86--92, November 2021.

\bibitem{ctc_2006}
Alex Graves, Santiago Fernández, Faustino Gomez, and Jürgen Schmidhuber.
\newblock Connectionist temporal classification: Labelling unsegmented sequence data with recurrent neural networks.
\newblock In {\em Proceedings of the 23rd International Conference on Machine Learning {(ICML)}}, volume 2006, pages 369--376, 01 2006.

\bibitem{rimes_2024}
E.~Grosicki, M.~Carré, E.~Geoffrois, E.~Augustin, F.~Preteux, and R.~Messina.
\newblock {RIMES}, complete.
\newblock \url{https://doi.org/10.5281/zenodo.10812725}, 2024.
\newblock [Last~Accessed:~20~May~2024].

\bibitem{rimes_2008}
Emmanu{\`e}le Grosicki, Matthieu Carre, Jean-Marie Brodin, and Edouard Geoffrois.
\newblock {RIMES evaluation campaign for handwritten mail processing}.
\newblock In {\em Proceedings of the 11th International Conference on Frontiers in Handwriting Recognition ({ICFHR})}, pages 1--6, Montreal, Canada, 2008. {Concordia University}.

\bibitem{gruendler:1993}
Beatrice Gruendler.
\newblock {\em The Development of the {A}rabic Scripts: From the {N}abatean Era to the First {I}slamic Century According to Dated Texts}.
\newblock Scholars Press, {A}tlanta, {GA}, 1993.

\bibitem{vml_hd_2017}
Majeed Kassis, Alaa Abdalhaleem, Ahmad Droby, Reem Alaasam, and Jihad El-Sana.
\newblock {VML-HD}: The historical {A}rabic documents dataset for recognition systems.
\newblock In {\em Proceedings of the 1st International Workshop on {A}rabic Script Analysis and Recognition ({ASAR})}, pages 11--14, 2017.

\bibitem{badam_2019}
Benjamin Kiessling, Daniel Stoekl Ben~Ezra, and Matthew Miller.
\newblock {BADAM}: A public dataset for baseline detection in {A}rabic-script manuscripts.
\newblock In {\em Proceedings of the 5th International Workshop on Historical Document Imaging and Processing ({HIP})}, pages 13--18, September 2019.

\bibitem{madcat_2012}
David Lee, Safa Ismael, Stephen Grimes, Dave Doermann, Stephanie Strassel, and Zhiyi Song.
\newblock {MADCAT Phase 1 Training Set LDC2012T15}.
\newblock \url{https://catalog.ldc.upenn.edu/LDC2012T15}, 2012.
\newblock [Last~Accessed:~18~April~2024].

\bibitem{levenshtein:1966}
Vladimir~I. Levenshtein.
\newblock Binary codes capable of correcting deletions, insertions, and reversals.
\newblock {\em Soviet physics. Doklady}, 10:707--710, 1965.

\bibitem{trocr_2021}
Minghao Li, Tengchao Lv, Chen Jingye, Lei Cui, Yijuan Lu, Dinei Florencio, Cha Zhang, Zhoujun Li, and Furu Wei.
\newblock {TrOCR}: Transformer-based optical character recognition with pre-trained models.
\newblock In {\em Proceedings of the 37th {AAAI} Conference on Artificial Intelligence}, 2021.

\bibitem{khatt_2012}
Sabri~A. Mahmoud, Irfan Ahmad, Mohammad Alshayeb, Wasfi~G. Al-Khatib, Mohammad~Tanvir Parvez, Gernot~A. Fink, Volker Märgner, and Haikal~El Abed.
\newblock {KHATT}: {A}rabic offline handwritten text database.
\newblock In {\em Proceedings of the International Conference on Frontiers in Handwriting Recognition ({ICFHR})}, pages 449--454, 2012.

\bibitem{iam_2002}
Urs-Viktor Marti and Horst Bunke.
\newblock The {IAM}-database: An {E}nglish sentence database for offline handwriting recognition.
\newblock {\em International Journal on Document Analysis and Recognition}, 5:39--46, November 2002.

\bibitem{ahtid_mw_2012}
Anis Mezghani, Slim Kanoun, Maher Khemakhem, and Haikal~El Abed.
\newblock A database for {A}rabic handwritten text image recognition and writer identification.
\newblock In {\em Proceedings of the International Conference on Frontiers in Handwriting Recognition ({ICFHR})}, pages 399--402, 2012.

\bibitem{arabictransformer_2024}
Saleh Momeni and Bagher BabaAli.
\newblock A transformer-based approach for {A}rabic offline handwritten text recognition.
\newblock {\em Signal, Image and Video Processing}, 18:3053--3062, 2024.

\bibitem{adocrnet_2024}
Lamia Mosbah, Ikram Moalla, Tarek~M. Hamdani, Bilel Neji, Taha Beyrouthy, and Adel~M. Alimi.
\newblock {ADOCRNet}: A deep learning {OCR} for {A}rabic documents recognition.
\newblock {\em IEEE Access}, 12:55620--55631, 2024.

\bibitem{ifn_farsi_2008}
Saeed Mozaffari, Haikal El~Abed, Volker Märgner, Karim Faez, and Seyed~Ali Amirshahi.
\newblock {IfN/Farsi-Database}: A database of {F}arsi handwritten city names.
\newblock In {\em Proceedings of the 11th International Conference on Frontiers in Handwriting Recognition ({ICFHR})}, Montreal, Canada, January 2008. {Concordia University}.

\bibitem{htr_dataset_2024}
Rayyan Najam and Safiullah Faizullah.
\newblock A scarce dataset for ancient {A}rabic handwritten text recognition.
\newblock {\em Data in Brief}, 56:2352--3409, 10 2024.

\bibitem{openai_chatgpt_api}
{OpenAI}.
\newblock {API Reference}.
\newblock Retrieved from \url{https://platform.openai.com/docs/api-reference/introduction}.
\newblock [Last~Accessed:~03~June~2024].

\bibitem{hadara_2014}
Werner Pantke, Martin Dennhardt, Daniel Fecker, Volker Märgner, and Tim Fingscheidt.
\newblock An historical handwritten {A}rabic dataset for segmentation-free word spotting --- {HADARA80P}.
\newblock In {\em Proceedings of the 14th International Conference on Frontiers in Handwriting Recognition ({ICFHR})}, pages 15--20, 2014.

\bibitem{hadara_2013}
Werner Pantke, Volker Märgner, Daniel Fecker, Tim Fingscheidt, Abedelkadir Asi, Ofer Biller, Jihad El-Sana, Raid Saabni, and Mohammad Yehia.
\newblock {HADARA} --- a software system for semi-automatic processing of historical handwritten {A}rabic documents.
\newblock In {\em Proceedings of the IS\&T Archiving Conference}, pages 161--166. Society for Imaging Science and Technology ({IS\&T}), 04 2013.

\bibitem{url-page-viewer}
{Pattern Recognition and Image Analysis Research Lab (PRImA)}.
\newblock {PAGE Viewer}.
\newblock \url{https://www.primaresearch.org/tools/PAGEViewer}.
\newblock [Last~Accessed:~20~May~2024].

\bibitem{url-page}
{Pattern Recognition and Image Analysis Research Lab (PRImA)}.
\newblock {PAGE XML} for page content.
\newblock \url{https://www.primaresearch.org/schema/PAGE/gts/pagecontent/2019-07-15/pagecontent.xsd}.
\newblock [Last~Accessed:~20~May~2024].

\bibitem{ifn_enit_2002}
Mario Pechwitz, Samia Snoussi, Volker Märgner, Noureddine Ellouze, and Hamid Amiri.
\newblock {IFN/ENIT}---{D}atabase of handwritten {A}rabic words.
\newblock In {\em 7th Colloque International Francophone sur l'Ecrit et le Document, {(CIFED)}}, October 2002.

\bibitem{page-2010}
Stefan Pletschacher and Apostolos Antonacopoulos.
\newblock The {PAGE} (page analysis and ground-truth elements) format framework.
\newblock In {\em Proceedings of the 20th International Conference on Pattern Recognition ({ICPR})}, pages 257--260, 2010.

\bibitem{icdar_2017}
Joan~Andreu Sánchez, Verónica Romero, Alejandro~H. Toselli, Mauricio Villegas, and Enrique Vidal.
\newblock {ICDAR2017} competition on handwritten text recognition on the read dataset.
\newblock In {\em Proceedings of the 14th {IAPR} International Conference on Document Analysis and Recognition {(ICDAR)}}, volume~01, pages 1383--1388, 2017.

\bibitem{rasam_2021}
Chahan Vidal-Gor{\`e}ne, No{\"e}mie Lucas, Cl{\'e}ment Salah, Ali{\'e}nor Decours-Perez, and Boris Dupin.
\newblock {RASAM} -- {A} dataset for the recognition and analysis of scripts in {A}rabic maghrebi.
\newblock In Elisa~H. Barney~Smith and Umapada Pal, editors, {\em Document Analysis and Recognition -- ICDAR 2021 Workshops}, pages 265--281, Cham, 2021. Springer International Publishing.

\bibitem{dan_2020}
Tianwei Wang, Yuanzhi Zhu, Lianwen Jin, Canjie Luo, Xiaoxue Chen, Yaqiang Wu, Qianying Wang, and Mingxiang Cai.
\newblock Decoupled attention network for text recognition.
\newblock In {\em Proceedings of the 34th {AAAI} Conference on Artificial Intelligence}, pages 12216--12224, 2020.

\bibitem{sfr_2018}
Curtis Wigington, Chris Tensmeyer, Brian Davis, William Barrett, Brian Price, and Scott Cohen.
\newblock Start, {F}ollow, {R}ead: End-to-end full-page handwriting recognition.
\newblock In {\em {Proceedings of the {E}uropean {C}onference on {C}omputer {V}ision {(ECCV)}}}, pages 372--388, 2018.

\bibitem{arabic_wikipedia}
Wikipedia.
\newblock List of countries and territories where {A}rabic is an official language.
\newblock \url{https://en.wikipedia.org/wiki/List_of_countries_and_territories_where_Arabic_is_an_official_language#cite_note-1}, 2024.
\newblock [Last~Accessed:~10~March~2024].

\end{thebibliography}
\appendix
\newcommand{\blue}{}
\section {JSON file format used in the Muharaf dataset} \label{appendix:json}
Each image in the Muharaf dataset comes with two corresponding JSON files. One file has the suffix ``annotate'' and the other has the suffix``tagged''. The two formats are almost identical, except for the addition of tags in the later one. From both JSON files, the transcription and polygonal coordinates of each line can be retrieved from the keys with the prefix ``line\_''. The tags of the ``page'' JSON format specify whether the marked polygonal area is a text line or region. Additional tags specify whether it is graphics, logo, letterhead, or stamp. The detailed JSON schema is given below: 
\begin{verbatim}
 {"$schema": "https://json-schema.org/draft/2020-12/schema",
  "title": "Page",
  "type": "object",
  "properties": {
    "transcriber": {
      "type": "string",
      "description": "Name of the transcriber of this page"
    },
    "taggingBy": {
      "type": "string",
      "description": "Name of the person who tagged page elements 
                     (if applicable)"
    },
    "writer": {
      "type": "string",
      "description": "Writer's name if known"
    },
    "comment": {
      "type": "string",
      "description": "Any comment on the page image"
    },
    { "type": "object",
      "patternProperties": {
      "^line_": {
        "type": "object",
        description: "Each polygonal line or region in page image. 
                      The tags specify its type"
        "properties": {
          "text": {
            "type": "string",
            "description": "Line transcription"
          },
          "coord": {
            "type": "array"
            "description": "Polygonal coordinates of text line as a flat 
                            array [x1, y1, x2, y2, ...]"
          },
          "tags": {
            "type": "object",
            "description": "Tags for additional page elements. 
                            Not required if only annotations and
                            transcriptions are needed."
            "properties": {
              "Region_paragraph": {
                "type": "integer",
                "enum": [0, 1]
                },
              "Region_heading": : {
                "type": "integer",
                "enum": [0, 1]
                },
              "Region_logo": : {
                "type": "integer",
                "enum": [0, 1]
                },
              "Region_letterhead": : {
                "type": "integer",
                "enum": [0, 1]
                },
              "Region_floating": : {
                "type": "integer",
                "enum": [0, 1]
                },
              "Region_pageNo": : {
                "type": "integer",
                "enum": [0, 1]
                },
              "Region_signature": : {
                "type": "integer",
                "enum": [0, 1]
                },
              "Region_graphic": : {
                "type": "integer",
                "enum": [0, 1]
                },
              "Region_decoration": {
                "type": "integer",
                "enum": [0, 1]
                },
              "Region_noise": : {
                "type": "integer",
                "enum": [0, 1]
                },
              "Region_separator": : {
                "type": "integer",
                "enum": [0, 1]
                },
              "Region_linedrawing": : {
                "type": "integer",
                "enum": [0, 1]
                },
              "Region_stamp": : {
                "type": "integer",
                "enum": [0, 1]
                },
              "Archivist_Tag": : {
                "type": "integer",
                "enum": [0, 1]
                },
              "Printed_Regular": : {
                "type": "integer",
                "enum": [0, 1]
                },
              "Printed_Italics": : {
                "type": "integer",
                "enum": [0, 1]
                },
              "Printed_Bold": : {
                "type": "integer",
                "enum": [0, 1]
                },
              "Orient_top_bottom": : {
                "type": "integer",
                "enum": [0, 1]
                },
              "Orient_bottom_top": : {
                "type": "integer",
                "enum": [0, 1]
                }
              }
            }
          }
        }
      }
    }
  } 
}
\end{verbatim}
\section{Collections in the Muharaf dataset} \label{appendix:collections}
The various collections included in the Muharaf dataset and their characteristics are listed in Table~\ref{table:collections}. Section~\ref{appendix:chatgpt} describes the procedure for generating the keywords for each collection. If a timeline was unavailable from the archival collection, we assigned it to a broad period. For example, for an individual's collection, we designated the timeline as spanning from the year of birth to the year of death. Future work can continue narrowing it down. The keywords were generated by using OpenAI's GPT 3.5 APIs~\cite{openai_chatgpt_api} (details in Section~\ref{appendix:chatgpt}).

    \newcolumntype{L}[1]{>{\raggedright\arraybackslash}p{#1}}
    \newcolumntype{C}[1]{>{\centering\arraybackslash}p{#1}}
    \newcolumntype{R}[1]{>{\raggedleft\arraybackslash}p{#1}}
    \begin{filecontents*}{csv/keyword_stats.csv}
No.|Prefix|Total Files|Name|Approximate Time Period|Keywords
1|00000|94|USEK, Phoenix Center, Patriarche S. Aouad and B. Massad Collection|1683--1756, 1806--1890 |church administration, legal, financial, official correspondence
2|158|35|USEK, Phoenix Center, Al Moutasarrifiya Collection|1861--1918|official correspondence, church administration, historical account, legal
3|159|5|USEK, Phoenix Center, Al Moutasarrifiya Collection|1861--1918|official correspondence, government, politics, legal
4|160|13|USEK, Phoenix Center, Al Moutasarrifiya Collection|1861--1918|official correspondence, politics, historical account, personal letter
5|2015|10|Khayrallah Center, El-Khouri Letters Collection||personal letter, church administration, historical account, official correspondence
6|0105-12P|11|USEK, Phoenix Center, PV Collection||church administration, financial, legal, historical account
7|1940Jbeil|19|USEK, Phoenix Center, PV Collection||church administration, financial, legal, historical account
8|26025|3|USEK, Phoenix Center, Al Moutasarrifiya Collection|1861--1918|legal, deed, official correspondence, marriage record
9|26066|3|USEK, Phoenix Center, Al Moutasarrifiya Collection|1861--1918|legal, financial, official correspondence, legal agreement
10|27404|15|USEK, Phoenix Center, Al Moutasarrifiya Collection|1861--1918|historical account, economy, financial, government
11|AF|25|USEK, Phoenix Center, Amin Farhat Collection|1916--1941|financial, personal letter, legal, official correspondence
12|AJN|13|USEK, Phoenix Center, Joseph Nehme Collection|1910--1994|historical account, personal letter, spiritual, biography
13|AP|29|USEK, Phoenix Center, Lebanese Maronite Missionaries Collection||church administration, legal, official correspondence, analysis
14|AR|159|Khayrallah Center, Ameen Rihani Collection|1876--1940|historical account, official correspondence, personal letter, biography
15|AnF|8|USEK, Phoenix Center, Antoine Ferjane Collection|1906--1931|church administration, historical account, official correspondence, personal letter
16|BEK|15|USEK, Phoenix Center, Bechara El Khoury Collection|1885--1968|poem, official correspondence, spiritual, legal
17|Baddour|3|Khayrallah Center. Baddour Collection||poem, spiritual, historical account, legal
18|Church|20|USEK, Phoenix Center, Churchrecord Collection||marriage record, church administration, baptism record, historical account
19|EAC|21|USEK, Phoenix Center, Elias Abou Shabake Collection|1926--1928, 1956--1968|poem, historical account, spiritual, biography
20|EAK|98|USEK, Phoenix Center, Bishop Abdallah Khoury Collection|1872--1949|church administration, official correspondence, personal letter, spiritual
21|ES|18|USEK, Phoenix Center, President Elias Sarkis Collection|1924--1985|government, historical account, politics, economy
22|FC|21|USEK, Phoenix Center, President Fouad Chehab Collection||historical account, personal letter, financial, biography
23|HM|41|USEK, Phoenix Center, Hanna Moussa Collection|1959--1970|poem, spiritual, historical account, biography
24|JEH|34|USEK, Phoenix Center, Joseph El Hachem Collection|1925--2018|historical account, poem, personal letter, spiritual
25|JPK|24|USEK, Phoenix Center, Jean Philipp Kmeid Collection||biography, poem, analysis, historical account
26|JoM|98|USEK, Phoenix Center, Joseph Mikhail Collection||church administration, legal, official correspondence, financial
27|KEllis|30|Khayrallah Center, Kellis Collection|1946--1987|personal letter, historical account, biography, official correspondence
28|KJoseph|22|Khayrallah Center, KJoseph Collection||official correspondence, financial, legal, historical account
29|ME|12|USEK, Phoenix Center, Mansour Eid Collection|1944--2013|analysis, politics, economy, historical account
30|MG|33|USEK, Phoenix Center, Maurice Gemayel Collection||government, official correspondence, economy, personal letter
31|MH|44|USEK, Phoenix Center, Father Michel Hayek Collection|1928--2005|historical account, spiritual, politics, analysis
32|MISC\_R|13|USEK, Phoenix Center, Five Registers Collection|1906--1911, 1953--1954|church administration, baptism record, marriage record, historical account
33|MK|25|USEK, Phoenix Center, Michel Kahwaji Collection|1912--2011|poem, biography, historical account, speech
34|MUKF|53|USEK, Phoenix Center, Papers of Kfarchima Municipality Collection||financial, official correspondence, legal, government
35|MeM|8|USEK, Phoenix Center, Meneem Meneem Collection|1837--1855, 1967|legal, church administration, biography, financial
36|Misc|25|USEK, Phoenix Center, Five Registers Collection|1906--1911, 1953--1954|baptism record, marriage record, church administration, historical account
37|Nasrallah|48|Khayrallah Center, Narallah Collection||historical account, biography, analysis, official correspondence
38|OLM|86|USEK, Phoenix Center, Five Registers Collection|1955--1963|legal, church administration, financial, deed
39|Oussani|21|Khayrallah Center. Oussani Collection (Ottomon)||historical account, travel log, government, official correspondence
40|QAJE|19|USEK, Phoenix Center, Jezzin Collection|1875--1950|legal, official correspondence, financial, historical account
41|QBat|40|USEK, Phoenix Center, Al Batroun Collection|1875--1950|legal, financial, official correspondence, deed
42|SB|13|USEK, Phoenix Center, Sami Behnan Collection||personal letter, spiritual, church administration, historical account
43|ST|48|USEK, Phoenix Center, Salah Tizani Collection|1959--2000|dialog, official correspondence, historical account, personal letter
44|TAttallah|30|Khayrallah Center, TAttallah Collection||official correspondence, personal letter, financial, historical account
45|TB|6|USEK, Phoenix Center, Toufic El Basha Collection|1924--2005|official correspondence, personal letter, biography, financial
46|YAL|21|USEK, Phoenix Center, Youssef Abdallah Lahoud Collection|1800--1924|personal letter, official correspondence, financial, legal
47|YFS|14|USEK, Phoenix Center, Youssef Fadlallah Salemeh Collection|1906--2001|personal letter, travel log, official correspondence, historical account
48|YS|81|USEK, Phoenix Center, Youssef Hanna El-Sawda Collection|1887--1969|historical account, official correspondence, politics, government
49|kc0061|49|Khayrallah Center, Miguel Saikali Collection||personal letter, historical account, travel log, spiritual
50|kc0066|68|Khayrallah Center, KC0066 Diary Collection||historical account, church administration, biography, travel log
\end{filecontents*}
\csvreader[separator=pipe,
        longtable={ L{0.5cm} L{1.5cm}  L{0.8cm}  L{2.9cm}  L{1.7cm} L{4cm}},
        table head= \caption{Collections in the Muharaf dataset. Each image's filename starts with the string shown in the ``Prefix'' column. \magentaText{The time period is listed if known.}}\label{table:collections}\\
        \hline 
        \toprule No.&Prefix & Total Files& Name& Approximate Time Period& Keywords\\ \midrule\endfirsthead

        \multicolumn{6}{l@{}}{\small Table~\ref{table:collections} continued\ldots}\\ \toprule
         No.&Prefix& Total Files& Name & Approximate Time Period& Keywords\\ \midrule\endhead
        
    \bottomrule\endfoot, 
        head to column names, 
        late after line=\\\addlinespace, 
        late after last line=\\
    ]{csv/keyword_stats.csv}{}{\csvlinetotablerow} 

\section{Procedure for generating keywords for each collection} \label{appendix:chatgpt}
We employed OpenAI's \texttt{gpt-3.5-turbo-0125} model~\cite{openai_chatgpt_api} to generate keywords for every collection in the Muharaf dataset. For each page image in a collection, we instructed GPT to generate a list of three keywords based on the Arabic text transcription of that page image. Next, we created a list of keywords and their corresponding frequencies for all pages in that collection. The top 4 occurring keywords present in the collection are listed in Table~\ref{table:collections}. The prompt used to generate keywords for an individual page image is given below:

\begin{verbatim}
You will be given Arabic text. Based on the text, assign it the most three 
relevant keywords from this list: 

1 official-correspondence
2 personal-letter
3 church-administration
4 analysis
5 legal
6 financial
7 marriage-record
8 baptism-record
9 census-record
10 economy
11 administrative
12 politics
13 speech
14 biography
15 government
16 poem
17 dialog
18 historical-account
19 commentary
20 spiritual
21 travel-log
22 deed
23 legal-agreement


Constraints:
1. Output the three keywords and their indices.
2. Keyword indices (1, 2) cannot occur together.
3. An index can only be a number from 1, 2, ...,  23.
4. Give output as JSON format with keys:
   keywords_list_of_length_3,indices_list_of_length_3
5. Don't output anything additional.
\end{verbatim}

Note that the prompt and the keywords present above were manually refined iteratively based on the output of GPT. \magentaText{We generated these keywords for the general interest of the research community who are non-Arabic speakers. However, we do not claim that these summaries and keywords are 100\% correct.}

\section{Downloading Muharaf data, related software, code, and license}\label{appendix:source_code}
The GitHub repository containing instructions on downloading the Muharaf dataset and links to all related code can be found at \url{https://github.com/mehreenmehreen/muharaf}. We briefly describe the contents of this repository and the associated licenses next. 
\subsection{Dataset download}
The Muharaf GitHub repository has a Zenodo link to download the public part of the Muharaf dataset. It has 1,216 images, which are hosted on Zenodo at \url{https://zenodo.org/records/11492215}. Users have the option to download the following:
\begin{itemize}
\item Public part of data files that contain page images and their corresponding annotation files. Both JSON and Page-XML files are included in the annotation files.
\item Individual line images extracted from the page images. The line images are available for the public part of the Muharaf dataset.
\item Summary and keyword files corresponding to each image of the public part of the Muharaf dataset. The keywords were extracted using the procedure described in Section\ref{appendix:chatgpt}.
\end{itemize}

The restricted part of the Muharaf dataset has 428 images distributed under a proprietary license. The images, annotations, line images, and summaries for the restricted portion of the Muharaf dataset can be obtained by writing to Carlos Younes~\texttt{carlosyounes@usek.edu.lb} at Phoenix Center for Lebanese Studies, USEK. 

\subsection{ScribeArabic annotation software}
All the annotations of the Muharaf dataset were created using the ScribeArabic software. The transcriptions for more than 1,400 lines were also entered using ScribeArabic. The ScribeArabic annotation software is a Django app. The source code for this app is hosted on GitHub at~\url{https://github.com/mehreenmehreen/ScribeArabic} and its manual is at~\url{https://github.com/mehreenmehreen/ScribeArabic/blob/main/manual.md}. While ScribeArabic was designed to annotate and transcribe Arabic page images, it can easily be adapted for other languages. It can also be adapted for labeling images for other computer vision applications.

\subsection{PAGE-XML converter and page elements viewer}\label{appendix:pageElements}
\begin{figure*}
\centering
\begin{subfigure}[t]{0.265\textwidth}
 \includegraphics[width=\linewidth]{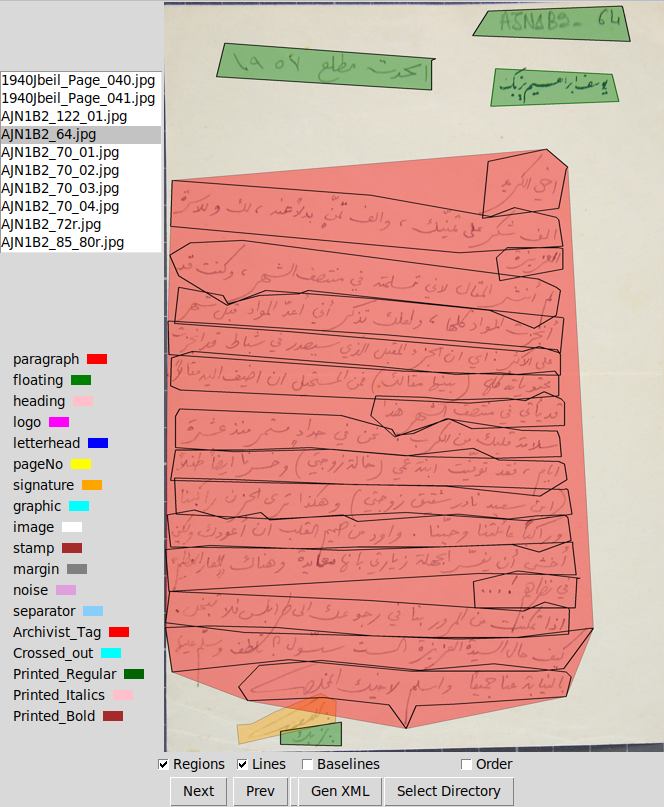}
    \caption{}
  \end{subfigure}
\hspace{0.06\textwidth}
\begin{subfigure}[t]{0.64\textwidth}
 \includegraphics[width=\linewidth]{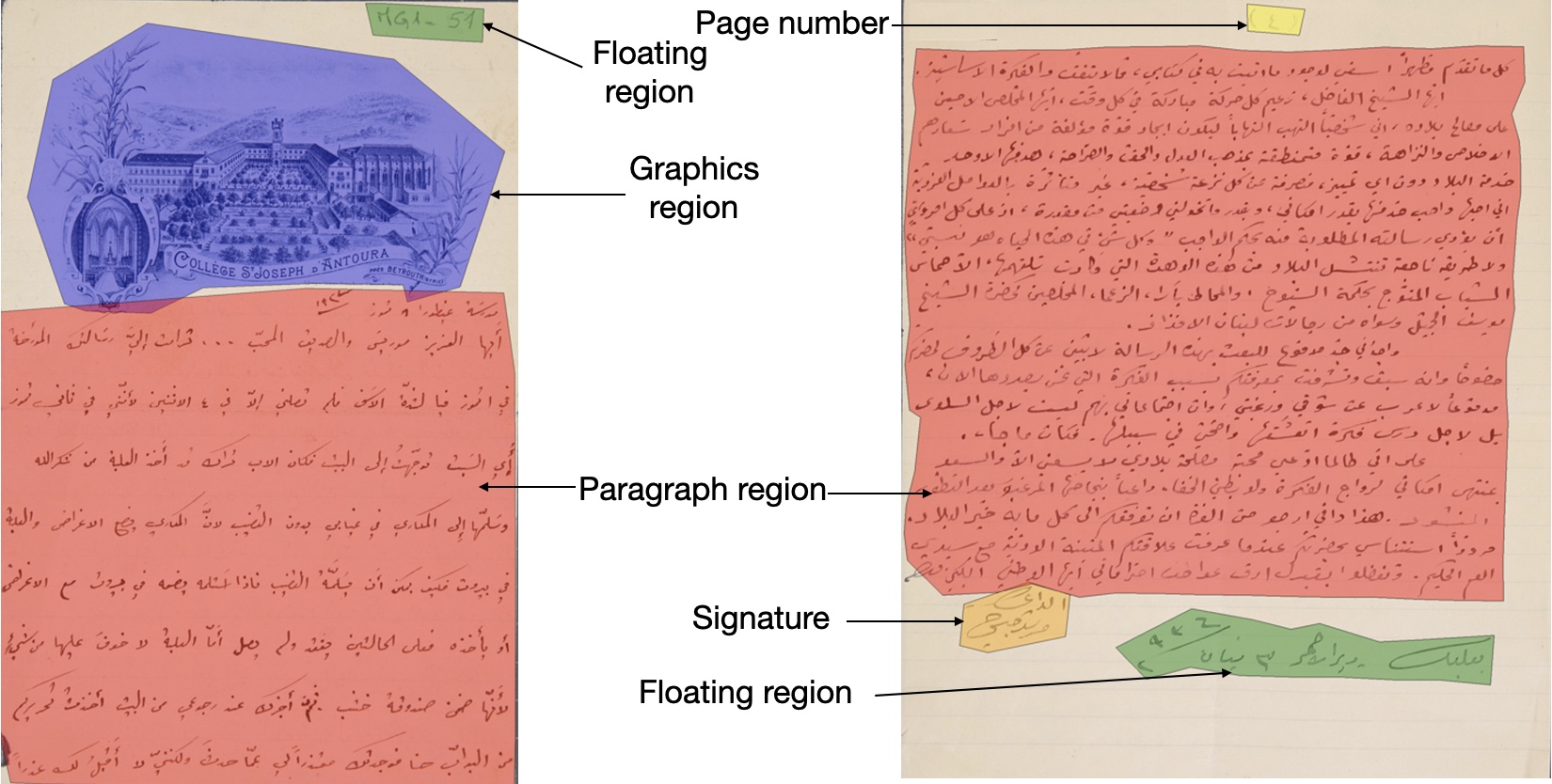}
    \caption{}
  \end{subfigure}

\caption{\magentaText{(a) Viewer of the annotated page elements showing floating regions, a paragraph region, and a signature region. Individual text lines are enclosed in black polygonal boundaries. (b) Various examples of different page elements including main paragraph regions, floating regions, a page number region, a signature region, and a graphics region.}}
\label{fig:pageElementsAll}
\end{figure*}

We include the code for converting ScribeArabic's JSON files to PAGE-XML files using a PAGE-XML converter. Instructions for downloading the code and running it are at \url{https://github.com/mehreenmehreen/xml_converter}. Users can also download a custom viewer for inspecting the annotated page elements on a document page. The viewer shows the different annotated page elements in different colors. A screenshot of this viewer is shown in Figure~\ref{fig:pageElementsAll}\magentaText{(a)}. This app was written in Python using the Tkinter library. \magentaText{Figure~\ref{fig:pageElementsAll}(b) illustrates examples of various page elements described in Section \ref{section:xml}.}



\subsection{Start, Follow, Read --- Arabic}
Users can replicate the results of all experiments reported in this paper by downloading the adapted source code of Start, Follow, Read (SFR)~\cite{sfr_2018}. The GitHub repository for the Arabic version is hosted at \url{https://github.com/mehreenmehreen/start_follow_read_arabic}. This repository contains code and links for the following:
\begin{itemize}
\item Code for preprocessing the image files and their corresponding JSON files. All data folders and files created after running the preprocessing step can also be downloaded via a provided Zenodo link.
\item Code for training the individual SOL, LF, and HW networks.
\item Folders for replicating the experiments on the public portion of the Muharaf dataset and the entire dataset. Both folders have three directories, \texttt{set0}, \texttt{set1}, and \texttt{set2}. Each set contains the training, validation, and test splits for running the experiments. Each set also contains the YAML configuration files that specify all the hyperparameters for training SFR-Arabic.
\item The trained SFR model weights obtained from training both the Muharaf dataset and its public portion. Users can run the inference code to get the error rates on various image files. Users can also run the full HTR code on a page image and generate JSON files containing the predictions from SFR-Arabic.
\item Additional inference results from training the Muharaf dataset and its public portion. \magentaText{There are also additional cross-dataset evaluation results.}
\end{itemize}
\subsection{\magentaText{Pretraining the model}}
\magentaText{The SOL and LF networks of SFR-Arabic were trained from scratch with random weights. We pretrained the HW model using synthetically generated Arabic lines. The text of the Arabic lines was taken from the Khaleej-2004 newspaper corpus~\cite{khaleej_2005} and Watan-2004 corpus~\cite{watan_2011}. We have provided the details of the pretrained model on our Github webpage with a link to download it.} 
\subsection{License}
We release our data and code under the following licenses:
\begin{itemize}
\item The public part of the Muharaf dataset has 1,216 images distributed using the Creative Commons license CC BY-NC-SA 4.0. Users are free to share and adapt the dataset under the terms of attribution, non-commercial use, and share alike, as specified by the Creative Commons license at \url{https://creativecommons.org/licenses/by-nc-sa/4.0/}. 
\item The restricted part of the dataset has 428 images distributed under a proprietary license. It can be downloaded by writing to Carlos Younes~\texttt{carlosyounes@usek.edu.lb} at Phoenix Center for Lebanese Studies, USEK. This part of the dataset is distributed under a proprietary license with the condition that it will not be redistributed and only be used for research purposes. 
\item The source code for the ScribeArabic annotation software and XML converter and viewer are also released under the Creative Commons license CC BY-NC-SA 4.0.  Users are free to share and adapt the dataset under the terms of attribution, non-commercial use, and share alike, as specified by the Creative Commons license at \url{https://creativecommons.org/licenses/by-nc-sa/4.0/}. 
\item \blue {The source code for \textsc{Start, Follow, Read --- Arabic} is modified from the source code released by the authors of Start, Follow, Read~\cite{sfr_2018}. Their code is free to use for academic and research purposes as given on their GitHub page: \url{https://github.com/cwig/start_follow_read?tab=readme-ov-file}. We release the Arabic version under the same license. \footnote{The authors agree to bear all responsibility in case of violation of rights, etc.}}
\end{itemize}

\subsection{\magentaText{Ethics statement: privacy, legal, or ethical issues}}
\magentaText{Muharaf-public has manuscript images from the archives of Phoenix Center for Lebanese Studies at USEK and KCLDS at NC State. These images were already publicly accessible before our proposed redistribution effort. Readers could access the images through the centers’ websites or as hard copies through the libraries. The research process described in this manuscript sticks to the ethical standards, and there are also no concerns regarding the leakage of personal information.}

\magentaText{Muharaf-restricted is distributed under a restricted license. The limited licensing is due to the proprietary nature of the images, and hence are authorized to use only with permission from the owners. The usage restrictions include that the dataset shall be used solely for research purposes and not be redistributed without permission. We will require researchers to agree to the statement of ethical use of data as a requirement of downloading the restricted part of the dataset.}

\section{\magentaText{Additional comparisons and cross-dataset validation results}}\label{appendix:comparison}
\magentaText{Table~\ref{table:IAM} shows the comparison of SFR with a transformer based system TrOCR~\cite{trocr_2021} on the IAM dataset~\cite{iam_2002}. While the performance of TrOCR is better than SFR, it has a higher computational complexity and requires more resources. Our choice of SFR is based on the goal of developing a system for a low-resource environment. Each network in this system can fit on an 8~GB card (mentioned in Section~\ref{section:evolution}), making this system ideal for a low-resource environment. State-of-the-art models like transformers cannot be realistically deployed in a resource-constrained setting. By choosing a more traditional CNN-based network, we are willing to trade a slight reduction in accuracy for a smaller, less resource-intensive model.} 

\magentaText{Table \ref{table:crossdataset} shows the results of cross-dataset evaluation on the KHATT dataset~\cite{khatt_2012}. When SFR was trained with Muharaf or RASAM+RASM datasets, the accuracy of both trained models was not very good. However, when all three datasets were combined, the system’s CER improved by~$\sim$16\%, showing the significance of Muharaf as a valuable addition to the current publicly available Arabic datasets. The first row of Table~\ref{table:crossdataset} shows the baseline CER of a transformer based system~\cite{arabictransformer_2024} trained on the KHATT dataset itself. We can see that the transformer has significantly large parameters~(153.1M) as compared to the HW network of SFR, which has only 18M parameters.}  

\begin{table}[htbp]
    \centering
    \caption{\magentaText{Comparison of TrOCR~\cite{trocr_2021} and SFR~\cite{sfr_2018} on the IAM dataset.}}   
    \label{table:IAM}
    \begin{filecontents*}{csv/IAM.csv}
Model|CER|Parameters
TrOCR\textsubscript{SMALL}  \cite{trocr_2021}|4.22|62M
TrOCR\textsubscript{BASE}  \cite{trocr_2021}|3.42|334M
TrOCR\textsubscript{LARGE} \cite{trocr_2021}|2.89|558M
SFR \cite{sfr_2018}||SOL-9M,
(used as baseline for Muharaf)|6.40|LF-5M, 
||HW-18M

\end{filecontents*}
\csvreader[separator=pipe,
        tabular={ L{4.5cm} C{2cm} C{1.5cm}  }, 
        table head=\toprule Model& CER (\%)&Parameters\\\midrule, 
        head to column names, 
        late after line=\\, 
        table foot = \bottomrule,
    ]{csv/IAM.csv}{} 
    {    \ifnum\thecsvrow=4\midrule\fi
    \csvcoli &\csvcolii &\csvcoliii
    }
\end{table}

\begin{table}[htbp]
    \centering
    \caption{\magentaText{HTR performance of systems on KHATT dataset trained with different datasets. The test images are the line images from the test set in KHATT’s ``unique paragraph'' directory.}}   
    \label{table:crossdataset}
    \begin{filecontents*}{csv/crossDataset.csv}
Model|Training Dataset|CER|Parameters
Transformer with cross-attention \cite{arabictransformer_2024}|KHATT|18.45|153.1M
HW (From SFR)|Muharaf|38.45|18M
HW (From SFR)|RASAM + RASM|41.14|18M
HW (From SFR)|Muharaf + RASAM + RASM|24.05|18M
\end{filecontents*}
\csvreader[separator=pipe,
        tabular={ L{5.5cm} C{3.5cm} C{1.2cm} C{1.3cm}}, 
        table head=\toprule Model& Training Dataset&CER (\%)&Parameters\\\midrule, 
        head to column names, 
        late after line=\\, 
        table foot = \bottomrule,
    ]{csv/crossDataset.csv}{} 
    {    \ifnum\thecsvrow=4\midrule\fi
    \csvcoli &\csvcolii &\csvcoliii&\csvcoliv
    }
\end{table}

\section{Datasheet for Muharaf Dataset}\label{appendix:datasheet}
We organize this section according to the relevant portions of the datasheet for datasets template~\cite{datasheet_2021}. 

\subsection{Motivation}
\paragraph {For what purpose was the dataset created?} \blue{The primary goal of creating this dataset is to train an OCR/HTR system capable of digitizing handwritten Arabic manuscripts and documents in a digital library, archive, or collection, making them accessible and searchable.} 

\paragraph{Who created the dataset?} \label{paragraph:qualifications} 
\magentaText{A majority of the annotations and transcriptions (1400+ images) were completed by Arabic speakers who manage digital archives of Arabic manuscripts and facilitate information access from a large number of Arabic documents. Their exact designations are:
\begin{itemize}
\item Carlos Younes: Head of the reference and external relations at Phoenix Center for Lebanese-Historical Archives (PCLS), USEK.
\item Georges Habchi: Head of Storage Division at Digital Development Center, USEK.
\end{itemize}
The transcriptions of both individuals were checked by a full professor of history who is also a historian and the director of USEK Library, Phoenix Center for Lebanese Studies.}

\magentaText{Some transcriptions (around 180 images) were completed by:
\begin{itemize}
\item Amin Elias: Assistant professor teaching history at the center for Lebanese studies.
\end{itemize}
These transcriptions were checked by a full professor of history at KCLDS, NC State. Each image of the packaged Muharaf dataset is accompanied by a JSON and XML file. The JSON and XML files contain the names of the annotator and the transcriber of that image. Table \ref{table:annotators} summarizes the total number of images transcribed by each team member.}
\paragraph{Who funded the creation of the dataset?} This work was supported in part by a grant from the National Endowment for the Humanities (NEH), FAIN: ZPA-283823-22. \blue{It was also supported in part by Family Search and the Electrical and Computer Engineering Undergraduate Research Program at NC State.}

\begin{table}[htbp]
    \centering
    \caption{\magentaText{Summary of total images transcribed by each member of the transcription team.}}   
    \label{table:annotators}
    \begin{filecontents*}{csv/annotators.csv}
Transcriber|Total Images
Amin Elias | 179
Carlos Younes | 663 
Georges Habchi | 802
Total | 1644
\end{filecontents*}
\csvreader[separator=pipe,
        tabular={ L{5.5cm} C{3.5cm}}, 
        table head=\toprule Transcriber& Total images transcribed\\\midrule, 
        head to column names, 
        late after line=\\, 
        table foot = \bottomrule,
    ]{csv/annotators.csv}{} 
    {    \ifnum\thecsvrow=4\midrule\fi
    \csvcoli &\csvcolii
    }
\end{table}

\subsection{Composition}
\paragraph{What do the instances that comprise the dataset represent?}
Each instance of the dataset is a document page image. Almost all images are scanned pages of handwritten Arabic, except 21 images that are in handwritten Ottoman Turkish. Three images are scanned typewritten pages. There are different types of individual page images, e.g., personal letters, poems, notes, diary images, legal correspondences, and church records. The timeline for this dataset ranges from the late 19th to the early 21st century.

\paragraph{How many instances are there in total?}
There are:\blue{
\begin{itemize}
\item \num{1644} image files of scanned handwritten Arabic document pages. \num{1216} image files are public and \num{428} files are restricted.
\item \num{36311} (\num{24495} public and \num{11816} restricted) text lines.
\end{itemize}
} 
\paragraph{Does the dataset contain all possible instances or is it a sample of instances from a larger set?}
\blue{The dataset is complete and contains all possible instances.}

\paragraph{What data does each instance consist of?}
\blue{Each instance consists of:
\begin{itemize}
\item A scanned handwritten Arabic document image (JPEG format). This is for researchers working on full page OCR/HTR.
\item A processed line image (PNG format). This is for researchers working on OCR/HTR of text line images.
\end{itemize}
}
\paragraph{Is there a label or target associated with each instance?}
The target/label for each page image represents:
\begin{itemize}
\item Spatial polygonal coordinates of individual text lines in a document image.
\item Transcription of each text line.
\item Spatial coordinates of a page element and its type. The type can be:
\begin{itemize}
\item Paragraph region. 
\item Floating text region. Any text outside the normal flow of text is labeled as a floating text region.
\item Page number region.
\item Signature region. This region contains names and signatures.
\item Graphics region. This region can contain logos, stamps, or letterhead images. Text lines are also allowed in this region.
\end{itemize}
 \end{itemize}
\blue{Each page image is accompanied by:
\begin{itemize}
\item One JSON file with ``\_annotate'' in its filename and containing the annotations and transcriptions of all text lines in the page image.
\item One JSON file with ``\_tagged'' suffix containing the annotation and transcriptions of each text line. It also contains the annotation of various page elements on the document image.
\item A PAGE-XML file for compatibility with PRImA Research Lab’s Aletheia tool~\cite{aletheia_2011} and PRImA Research Lab's PAGE-XML viewer~\cite{url-page-viewer}.
\item A plain text file containing an English summary of the ground truth of the document page image.
\item A plain text file containing keywords of the document page image in English. The keywords were generated using OpenAI's GPT APIs~\cite{openai_chatgpt_api}. 
\end{itemize}
For each line image, the label is its transcription, which is contained in a plain text file. 
}
\paragraph{Is any information missing from individual instances or labels?
}
A few text lines in the margins, footnotes, or signatures are not annotated and transcribed. Also, some of the page elements like separators or noise are not annotated.

\paragraph{Are there recommended data splits?}
No. We trained the system with three different random splits of training, validation, and test sets (1500, 50, 96). Our GitHub website includes links to download the three different sets of data splits.

\paragraph {Are there any errors, sources of noise, or redundancies in the dataset?}
As images vary in each document, the labeling of page elements such as floating text, heading, and main paragraphs was done using the personal judgment of the annotator. While the transcriptions have gone through a QA round, there are still some minor errors in the dataset:
\begin{itemize}
\item In very rare cases, some words may not be transcribed properly.
\item The annotations of many text lines were generated automatically, and hence, they are not tight polygons around the line. This implies that many annotations overlap and a bounding polygon may contain some portions of the polygon from the line above or below it.
\item Some signatures and floating areas have multiple text lines contained in the same line annotation. 
\end{itemize}

\paragraph{Is the dataset self-contained, or does it link to or otherwise rely on external resources?}
The dataset is self-contained.

\paragraph{Does the dataset contain data that might be considered confidential?}
The dataset does not contain any confidential information. The public portion of the dataset includes page images that are already publicly accessible. For the restricted part of the dataset, the images are proprietary and require permission from the owner for use only in non-commercial research.

\paragraph{Does the dataset relate to people?}
Yes. The dataset includes many document images such as personal letters, diaries, and official records from various personal collections. The documents are from the early 19th century to the early 21st century and do not relate to living people.

\paragraph{Does the dataset identify any subpopulations?}
No

\paragraph{Is it possible to identify individuals, either directly or indirectly from the dataset?}
Yes, for many page images, the writer of the document can be identified.

\paragraph{Does the dataset contain data that might be considered sensitive in any way?}
No

\subsection{Collection process}
\paragraph{How was the data associated with each instance acquired?}
An image from a particular archive collection was identified and used for annotation and transcription.

\paragraph{ What mechanisms or procedures were used to collect the data?}
\begin{itemize}
\item For 180 images, the text lines were annotated using the ScribeArabic software. The text was entered in an Excel sheet by a history professor and manually verified by another history professor.
\item During the data collection process, the available data was used to train the Start, Follow, Read (SFR) system~\cite{sfr_2018}. The annotations and transcriptions of many images were automatically generated using SFR. 
\item For more than \num{1400} images, a team of two Lebanese Arabic speakers who are also archivists used the ScribeArabic software to either: 
\begin{itemize}
    \item Correct the annotations and transcriptions generated by SFR.
    \item Manually annotate and transcribe the text lines on a document page image from scratch.
\end{itemize}
\item The annotations and transcriptions of the \num{1400}+ images were checked by a native Lebanese Arabic expert who is also a historian. 
\item A team of non-Arabic speakers annotated the images with various page elements. These page elements were also manually verified.
\end{itemize}
\paragraph{Effectiveness of QA process} \magentaText{To assess the effectiveness of the QA process, we evaluated the CER and WER of several batches of received transcriptions both before and after the QA process. The results are summarized in Table \ref{table:QA}. As shown, only minimal changes were made to the transcriptions during the QA phase.}
\paragraph{Over what timeframe was the data collected?} March 2023--April 2023, July 2023--March 2024.
\paragraph{Were any ethical review processes conducted?}
No

\begin{table}[htbp]
    \centering
    \caption{\magentaText{CER and WER of transcriptions of 5 different batches before and after the QA phase.}}   
    \label{table:QA}
    \begin{filecontents*}{csv/qa.csv}
Batch Receiving Date| CER | WER | Total Files
12/01/2023 | 0.051 | 0.197|53
12/23/2023 | 0.113 | 0.581|55
01/02/2024 | 0.256 | 0.890|83
01/16/2024 | 0.176 | 0.700|98
01/26/2024 | 0.071 | 0.397|67
\end{filecontents*}
\csvreader[separator=pipe,
        tabular={ L{2.5cm} L{1.5cm}L{1.5cm}C{1.5cm}}, 
        table head=\toprule Date Batch Received (2024)& CER (\%) & WER (\%)&Total Images\\\midrule, 
        head to column names, 
        late after line=\\\addlinespace, 
        table foot = \bottomrule,
    ]{csv/qa.csv}{} 
    {    
    \csvcoli &\csvcolii&\csvcoliii&\csvcoliv
    }
\end{table}

\subsection{ Preprocessing/cleaning/labeling}

\paragraph{ Was any preprocessing/cleaning/labeling of the data done?}Yes.
\begin{itemize}
\item Some page images were cropped before annotation.
\item The folder with line images has text lines along with their corresponding transcriptions. The individual lines were extracted from document images using preprocessing routines of the SFR system.
\end{itemize}
\paragraph{Is the software used to preprocess/clean/label the instances available?}
Yes, the software is available.
\begin{itemize}
\item The original SFR code is available at: \url{https://github.com/cwig/start_follow_read}
\item The SFR code adapted for Arabic is available at \url{https://github.com/mehreenmehreen/muharaf} link.
\end{itemize}

\subsection{Uses}
\paragraph{Has the dataset been used for any tasks already?}
\blue{Yes, it has been used to train a full-page HTR system as described in Section \ref{section:sfr} of the main paper.} 
\paragraph{Is there a repository that links to any or all papers or systems that use the dataset?}
We have set up a GitHub page for this purpose \url{https://github.com/mehreenmehreen/muharaf}. 

\paragraph{What tasks could the dataset be used for?}
The dataset can be used to:
\begin{itemize}
\item Train a system for text line segmentation. 
\item Train a system for OCR/HTR.
\item Train a system for layout detection.
\item Train a language model using the ground truth transcriptions.
\item Linguists can study the colloquial form of Arabic for various periods.
\item Identify various writing styles in a given period. 
\item Identify various writing styles used to record legal documents or church records.
\end{itemize}

\paragraph{Is there anything about the composition of the dataset or the way it was collected and preprocessed/cleaned/labeled that might impact future uses?} No.
\subsection{Distribution}

\paragraph{Will the dataset be distributed to third parties outside of the entity on behalf of which the dataset was created?}
Yes, the dataset will be distributed to researchers.

\paragraph{How will the dataset be distributed?}
The public portion of the Muharaf dataset is hosted on Zenodo at \url{https://zenodo.org/records/11492215}. The related software and source code is hosted on GitHub at \url{https://github.com/mehreenmehreen/muharaf}. This GitHub page includes instructions on downloading the restricted portion of Muharaf.

\paragraph{When will the dataset be distributed?}
The links for downloading Muharaf-public are active now. The restricted part of Muharaf is also available upon request.

\paragraph{Will the dataset be distributed under a copyright or other intellectual property (IP) license, and/or under applicable terms of use (ToU)?}
\begin{itemize}
\item The public part of the dataset has \num{1216} images distributed using the Creative Commons license CC BY-NC-SA 4.0.  Users are free to share and adapt under the attribution, non-commercial, and share alike terms of the Creative Commons license as given at \url{https://creativecommons.org/licenses/by-nc-sa/4.0/}. 
\item The restricted part of the dataset has \num{428} images distributed under a proprietary license. It can be downloaded by writing to Carlos Younes~\texttt{carlosyounes@usek.edu.lb} at Phoenix Center for Lebanese Studies, USEK. This part of the dataset is distributed under a proprietary license with the condition that it will not be redistributed and only be used for research purposes. 
\end{itemize}
\paragraph{Have any third parties imposed IP-based or other restrictions on the data associated with the instances?}
The restricted part of the dataset has \num{428} images distributed under a proprietary license with permission from the owners. 
\paragraph{Do any export controls or other regulatory restrictions apply to the dataset or to individual instances?} No.
\subsection{Maintenance}

\paragraph{Who is supporting/hosting/maintaining the dataset?}
The dataset is hosted on Zenodo. It will be maintained by the research teams working at Khayrallah Center for Lebanese Diaspora Studies 
(KCLDS) and Electrical and Computer Engineering Department (ECE) at NC State. 

\paragraph{How can the owner/curator/manager of the dataset be contacted?}
Users can contact the teams managing the dataset by directly opening an issue on the GitHub page: \url{ https://github.com/mehreenmehreen/muharaf}. They can also contact them using the emails provided on the GitHub page.

\paragraph{Is there an erratum?}
We'll build an erratum over time as more and more researchers start using Muharaf. 

\paragraph{Will the dataset be updated?}
Yes, transcription errors pointed out by users will be corrected and updated after verification.

\paragraph {Will older versions of the dataset continue to be supported/hosted/maintained?}
Yes, we plan to host all versions of our dataset on Zenodo.

\paragraph{If others want to extend/augment/build on/contribute to the dataset, is there a mechanism for them to do so?}
We have released the code for the ScribeArabic software, which can be used to annotate and transcribe page images. It can also be used for labeling various page elements. Other researchers can use ScribeArabic to build similar datasets. We welcome any additional contributions and are willing to add them to our dataset after verification.

\end{document}